%% file: ms.tex
\RequirePackage{snapshot}
\documentclass[10pt,twocolumn,letterpaper]{article}

\usepackage{cvpr}
\usepackage{times}
\usepackage{epsfig}
\usepackage{graphicx}
\usepackage{amsmath}
\usepackage{amssymb}

\usepackage{subcaption} %
\usepackage{siunitx}
\usepackage{booktabs} %
\usepackage{placeins} %
\usepackage{mathtools} %

\usepackage[pagebackref=true,breaklinks=true,letterpaper=true,colorlinks,bookmarks=false]{hyperref}

\cvprfinalcopy %

\def\cvprPaperID{5718} %

\ifcvprfinal\fi

\newcommand{\citep}{\cite} %

\newcommand{\tz}{\tilde{z}}
\newcommand{\tail}{G}
\newcommand{\RR}{\mathbb{R}}

\newcommand{\defeq}{\mathrel{\vcenter{\baselineskip0.5ex \lineskiplimit0pt
\hbox{\scriptsize.}\hbox{\scriptsize.}}}%
=}

\newcommand\blfootnote[1]{%
  \begingroup
  \renewcommand\thefootnote{}\footnote{#1}%
  \addtocounter{footnote}{-1}%
  \endgroup
}

\newcommand{\pictogram}{
\begin{figure}[t]
\centering
\includegraphics[width=0.325\textwidth]{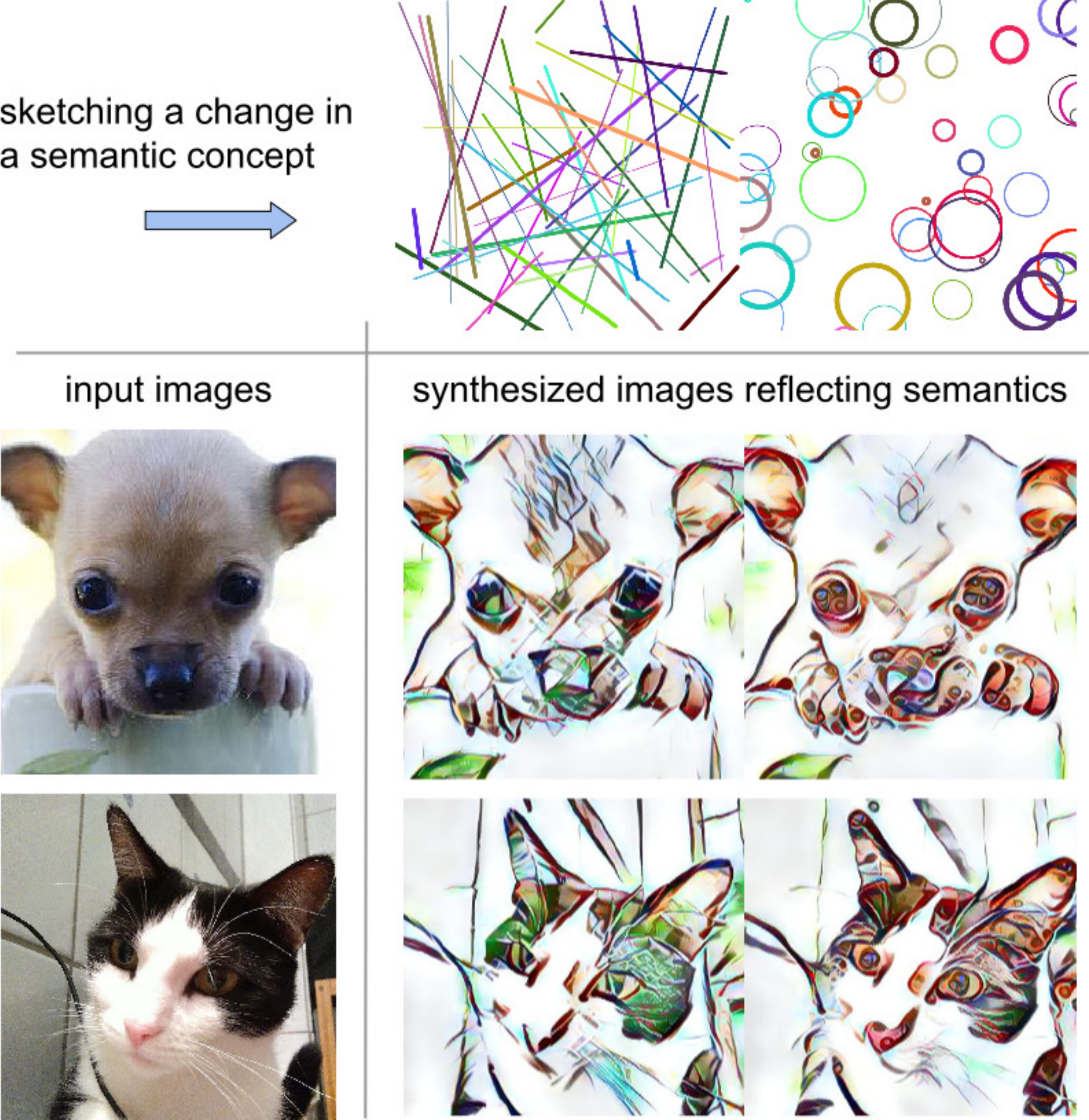}
\caption{Efficient generation of training examples for semantic concepts: A user must only provide
  two sketches (first row) for a change of a semantic concept, here: \emph{roundness}.
We then synthesize training images to reflect this semantic change.  
  }
  \label{fig:pictogram} \end{figure}
}
\newcommand{\celebainterpolation}{
\begin{figure}[b]
\centering
\includegraphics[width=0.305\textwidth]{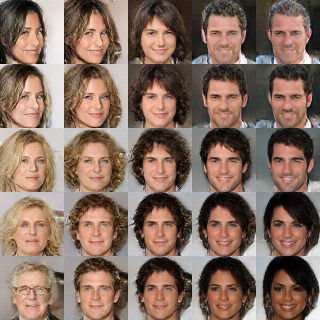}
\caption{CelebA: Four randomly drawn samples (corners) and corresponding
  interpolations obtained with unsupervised training, see
  Sec.~\ref{sec:aeint}.}
  \label{fig:celebainterpolation}
\end{figure}
}

\newcommand{\celebaclusterinterpolations}{
\begin{figure}[b]
\centering
\includegraphics[width=0.290\textwidth]{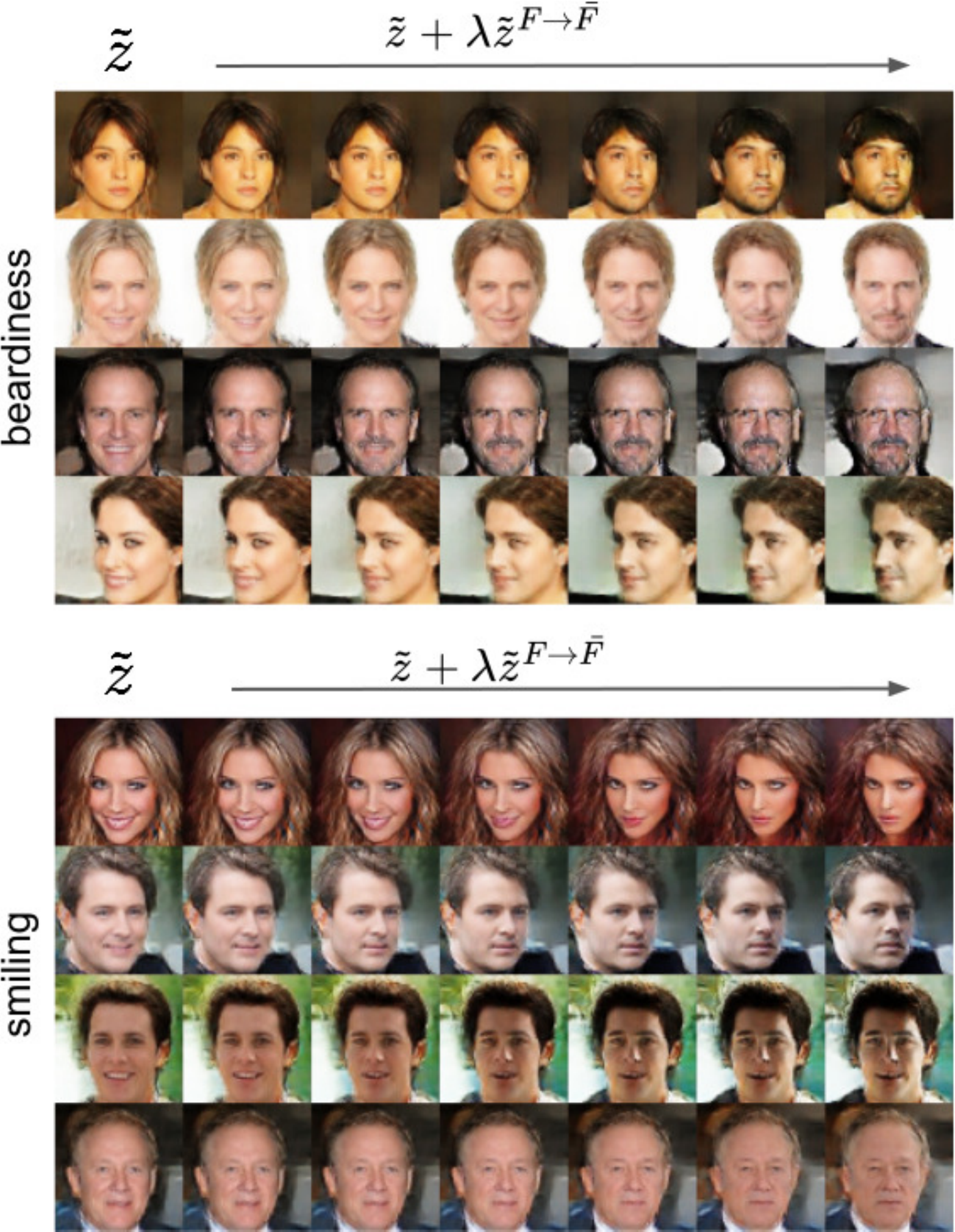}
\caption{Interpolating along semantic directions in disentangled representation space: First four rows show interpolations along \textsl{beardiness}, while the last four depict interpolations along \textsl{smiling} attribute. Note the change of gender in rows 1,2,4, reflecting the strong correlation of \textsl{beard} and \textsl{gender} in the original data.}
\label{fig:celebacluster}
\end{figure}
}

\newcommand{\celebabothstagesamples}{
\begin{figure*}[!htb]
\centering
\begin{minipage}{0.75\textwidth}
\includegraphics[width=\textwidth]{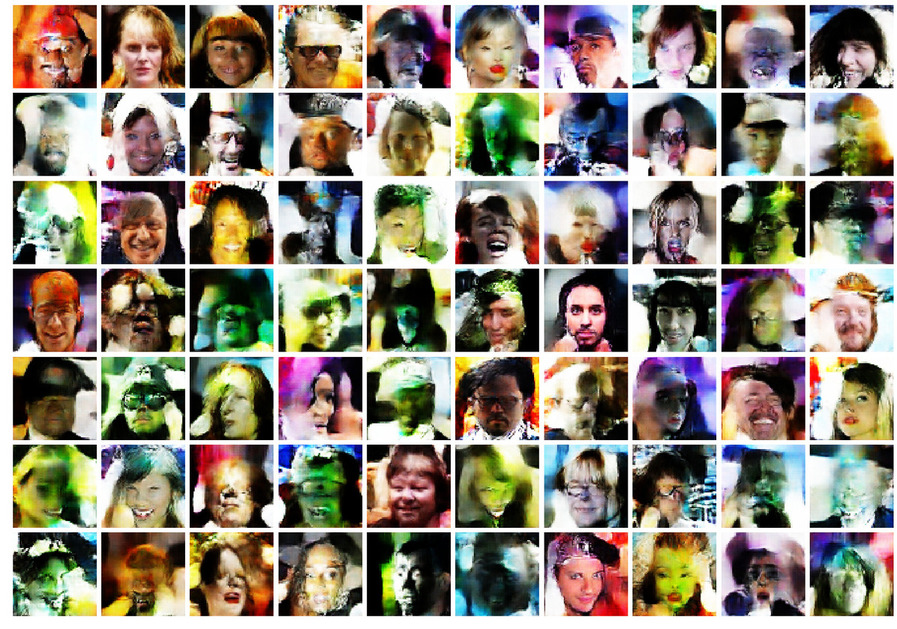}
\end{minipage}
\begin{minipage}{0.75\textwidth}
\includegraphics[width=\textwidth]{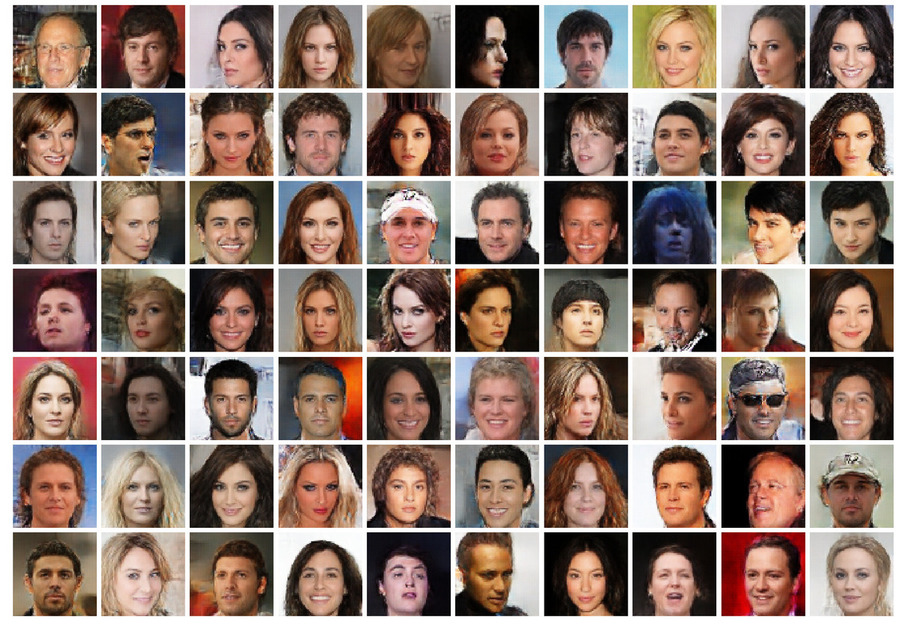}
\end{minipage}
\caption{Samples on CelebA. Top: Decoded samples drawn from the prior of the
  autoencoder: $x=G(z), \; z \sim \mathcal{N}(0, \mathbf{1})$. Bottom:
  Decoded samples drawn from the prior of the transformer:
  $x=G(T^{-1}(\tz)),  \; \tz \sim \mathcal{N}(0, \mathbf{1})$.}
  \label{fig:celebabothstagesamples}
\end{figure*}
}

\newcommand{\fashionmnistbothstagesamples}{
\begin{figure*}[!htb]
\centering
\begin{minipage}{0.475\textwidth}
\includegraphics[width=\textwidth]{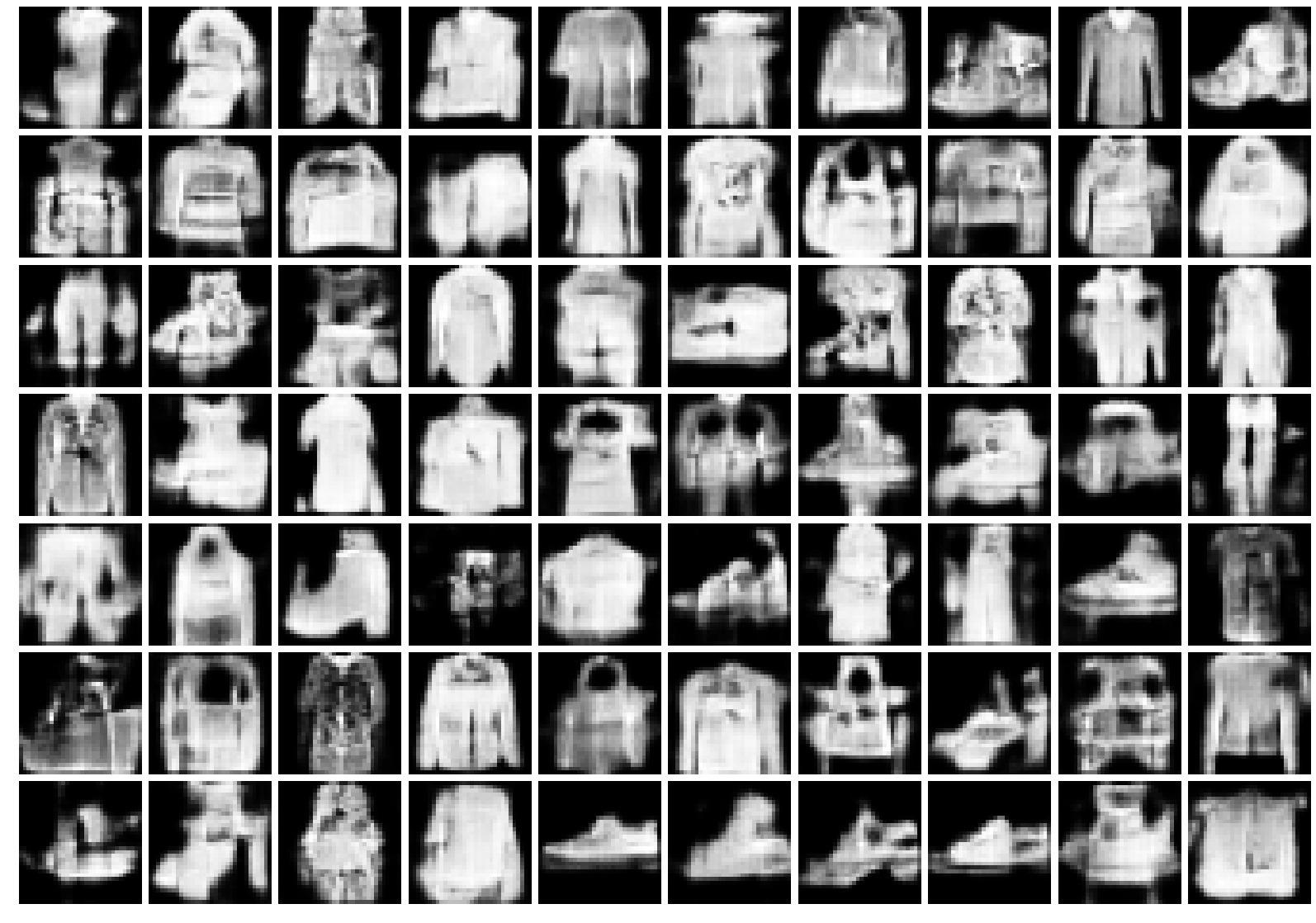}
\end{minipage}
\begin{minipage}{0.475\textwidth}
\includegraphics[width=\textwidth]{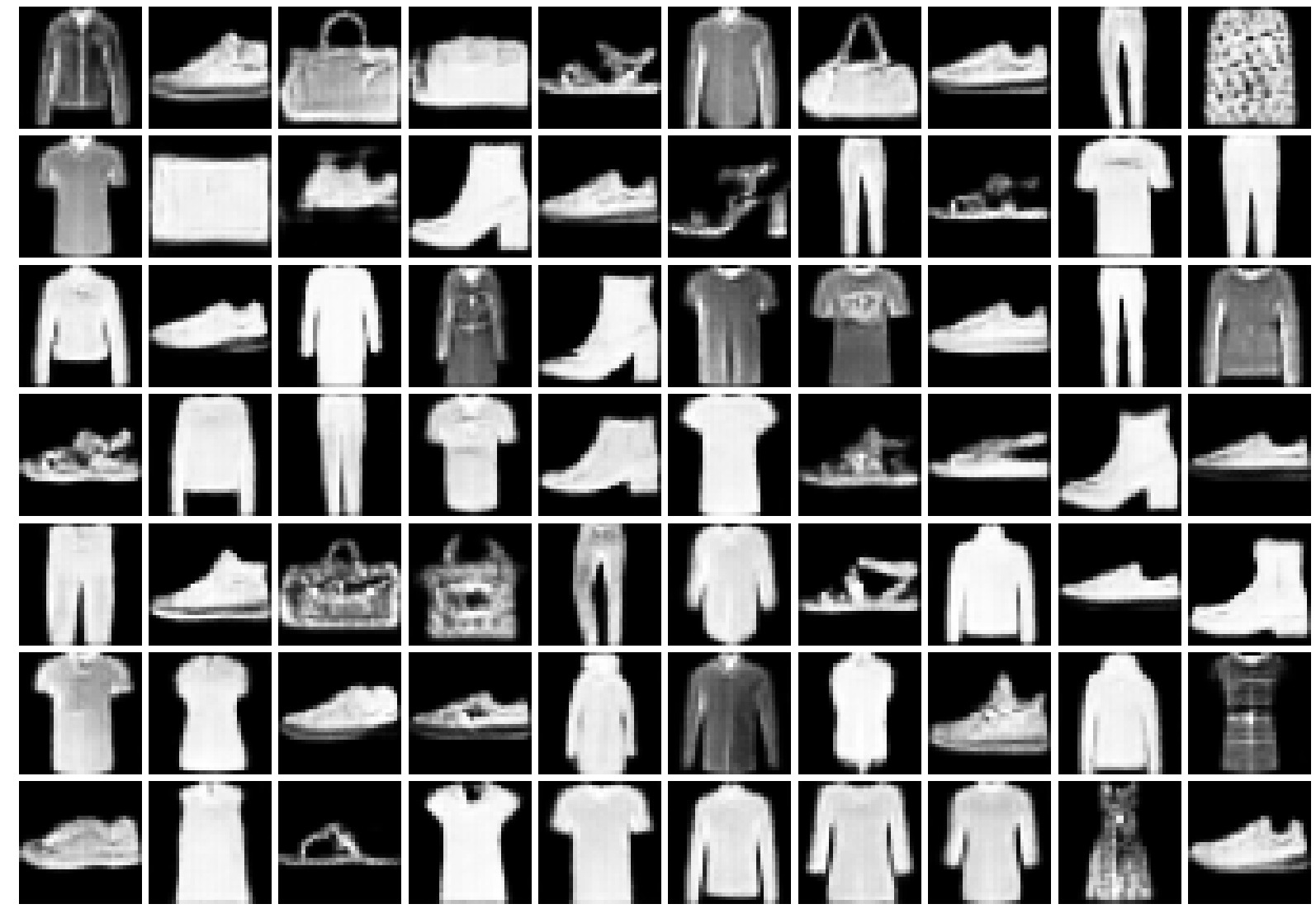}
\end{minipage}
\caption{Samples on FashionMNIST. Left: Decoded samples drawn from the prior
  of the autoencoder: $x=G(z), \; z \sim \mathcal{N}(0, \mathbf{1})$.
  Right: Decoded samples drawn from the prior of the transformer:
  $x=G(T^{-1}(\tz)), \; \tz \sim \mathcal{N}(0, \mathbf{1})$.}
  \label{fig:fashionmnistbothstagesamples}
\end{figure*}
}

\newcommand{\cifarbothstagesamples}{
\begin{figure*}[!htb]
\centering
\begin{minipage}{0.475\textwidth}
\includegraphics[width=\textwidth]{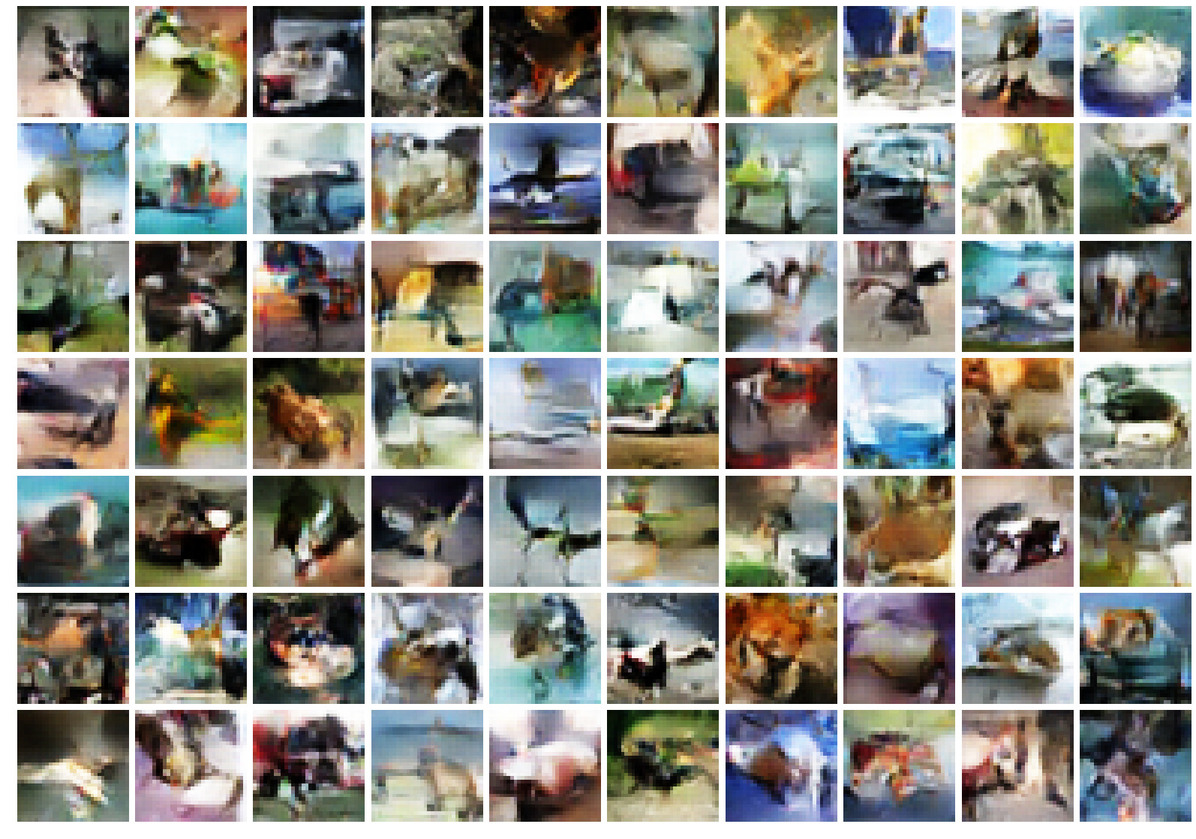}
\end{minipage}
\begin{minipage}{0.475\textwidth}
\includegraphics[width=\textwidth]{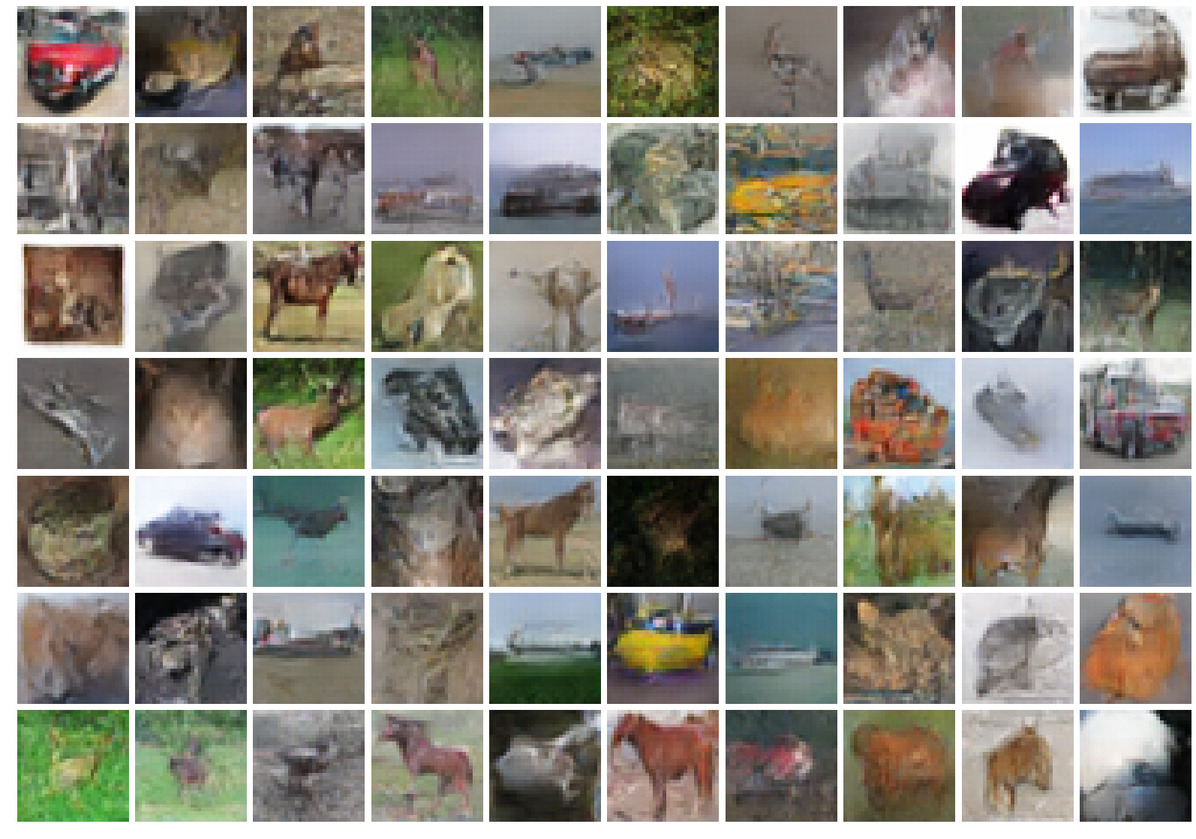}
\end{minipage}
\caption{Samples on CIFAR-10. Left: Decoded samples drawn from the prior of
  the autoencoder: $x=G(z), \; z \sim \mathcal{N}(0, \mathbf{1})$. Right:
  Decoded samples drawn from the prior of the transformer:
  $x=G(T^{-1}(\tz)),  \; \tz \sim \mathcal{N}(0, \mathbf{1})$.}
  \label{fig:cifarbothstagesamples}
\end{figure*}
}

\newcommand{\overview}{
\begin{figure}[t]
\centering
\includegraphics[width=0.405\textwidth]{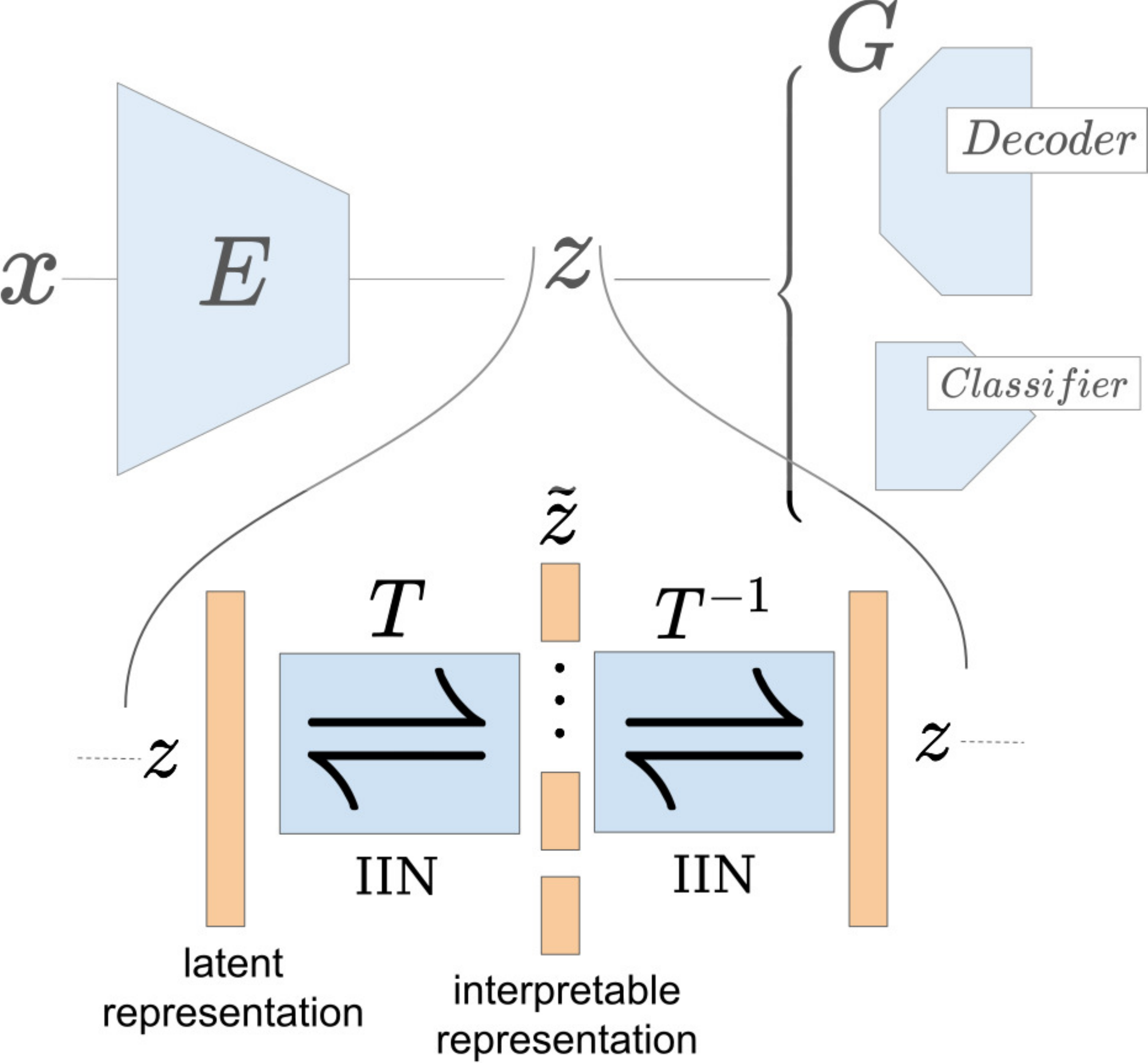}
\caption{Our Invertible Interpretation Network $T$ can be applied to
  arbitrary existing models. Its invertibility guarantees that the
  translation from $z$ to $\tz$ does not affect the performance of the model
  to be interpreted. Code and results can be found at the project page
  \href{https://compvis.github.io/iin/}{https://compvis.github.io/iin/}.
  }
  \label{fig:overview}
\end{figure}
}

\newcommand{\transformerarch}{
\begin{figure}[!htb]
\centering
\includegraphics[width=0.475\textwidth]{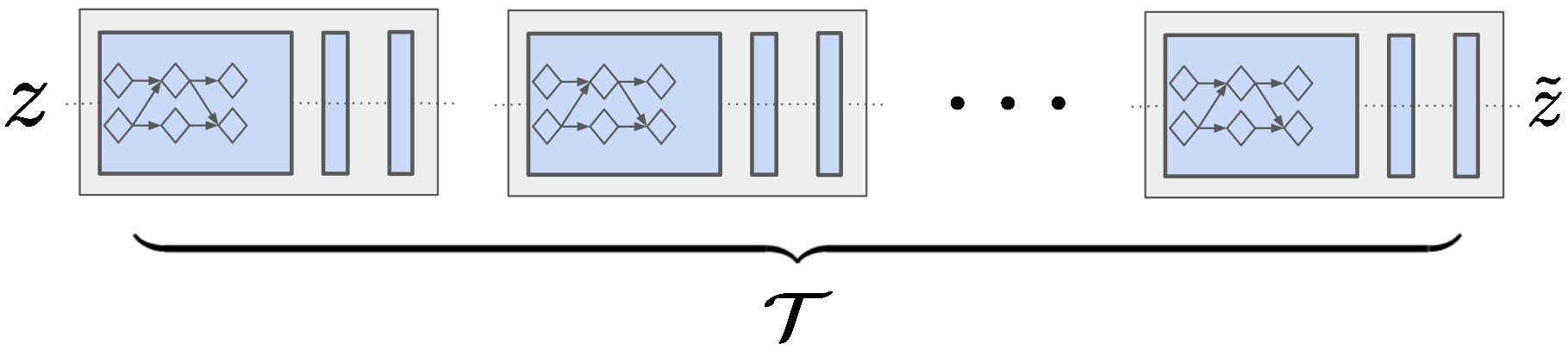}
\caption{Overview of our invertible interpretation network consisting of
  multiple bocks, see Fig.~\ref{fig:flowblock}.}
\label{fig:transformerarch}
\end{figure}
}

\newcommand{\flowblock}{
\begin{figure}[!htb]
\centering
\includegraphics[width=0.285\textwidth]{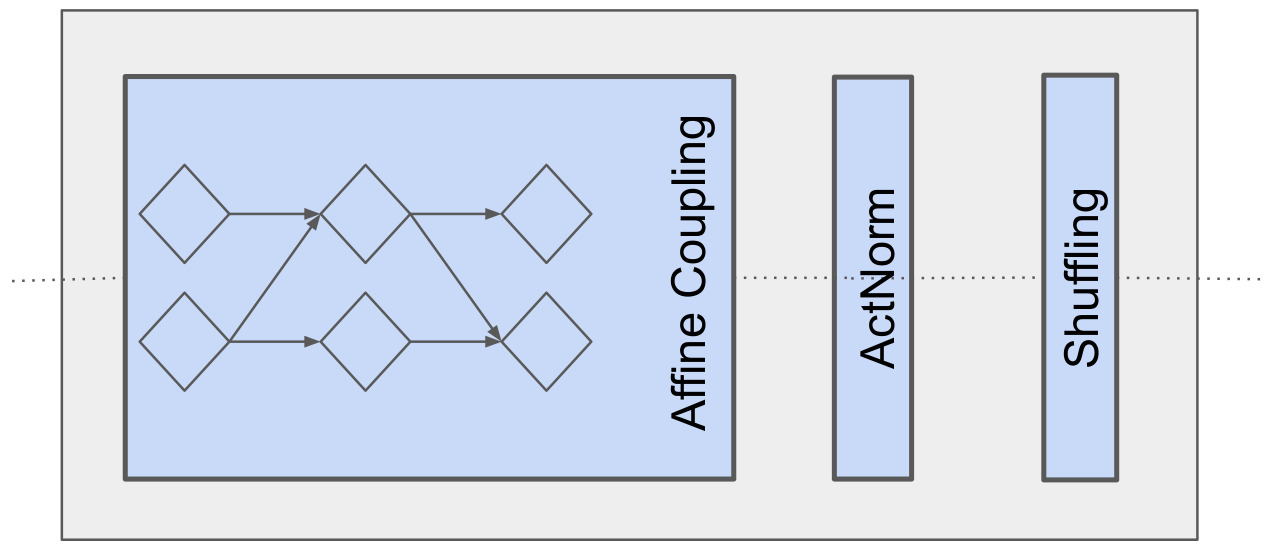}
\caption{A single block of our interpretation network built from invertible layers described
  in Sec.~\ref{sec:iinarchitecture}.}
  \label{fig:flowblock}
\end{figure}
}

\newcommand{\cmnistresponse}{
\begin{figure*}[t]
\centering
    \includegraphics[width=0.33\textwidth]{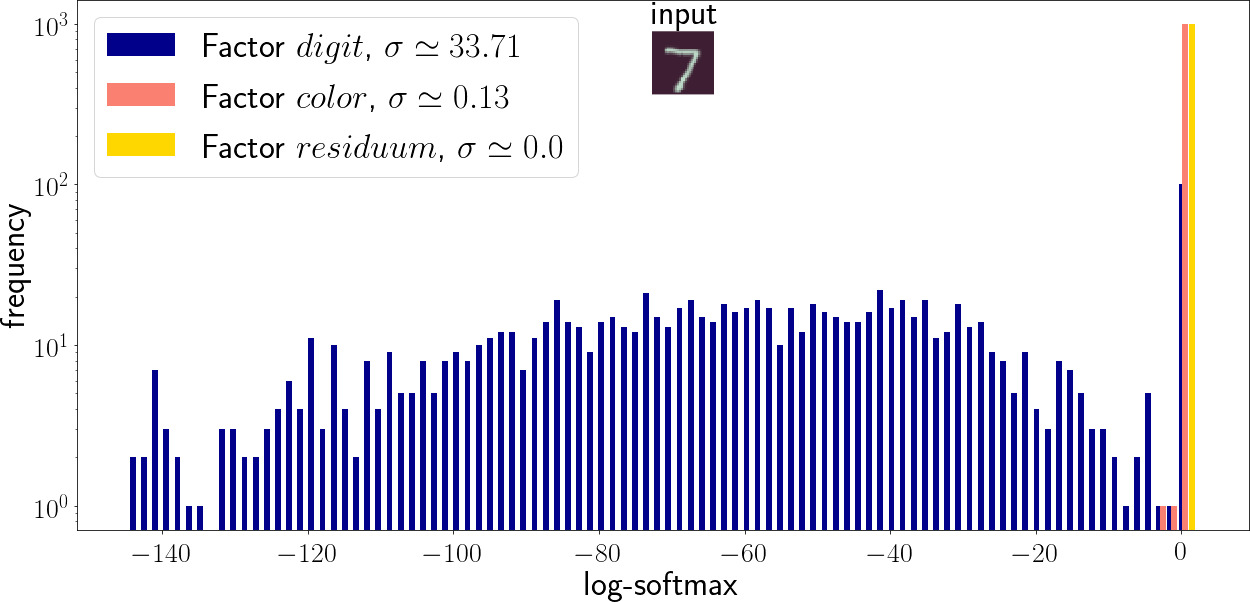}
    \includegraphics[width=0.33\textwidth]{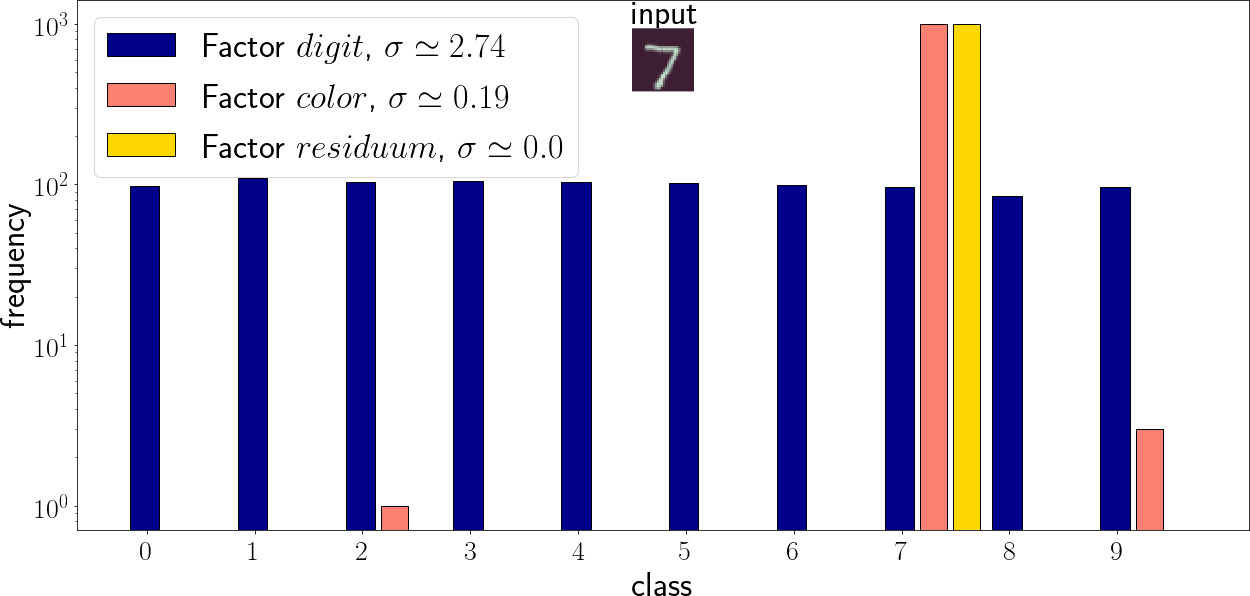}
    \includegraphics[width=0.33\textwidth]{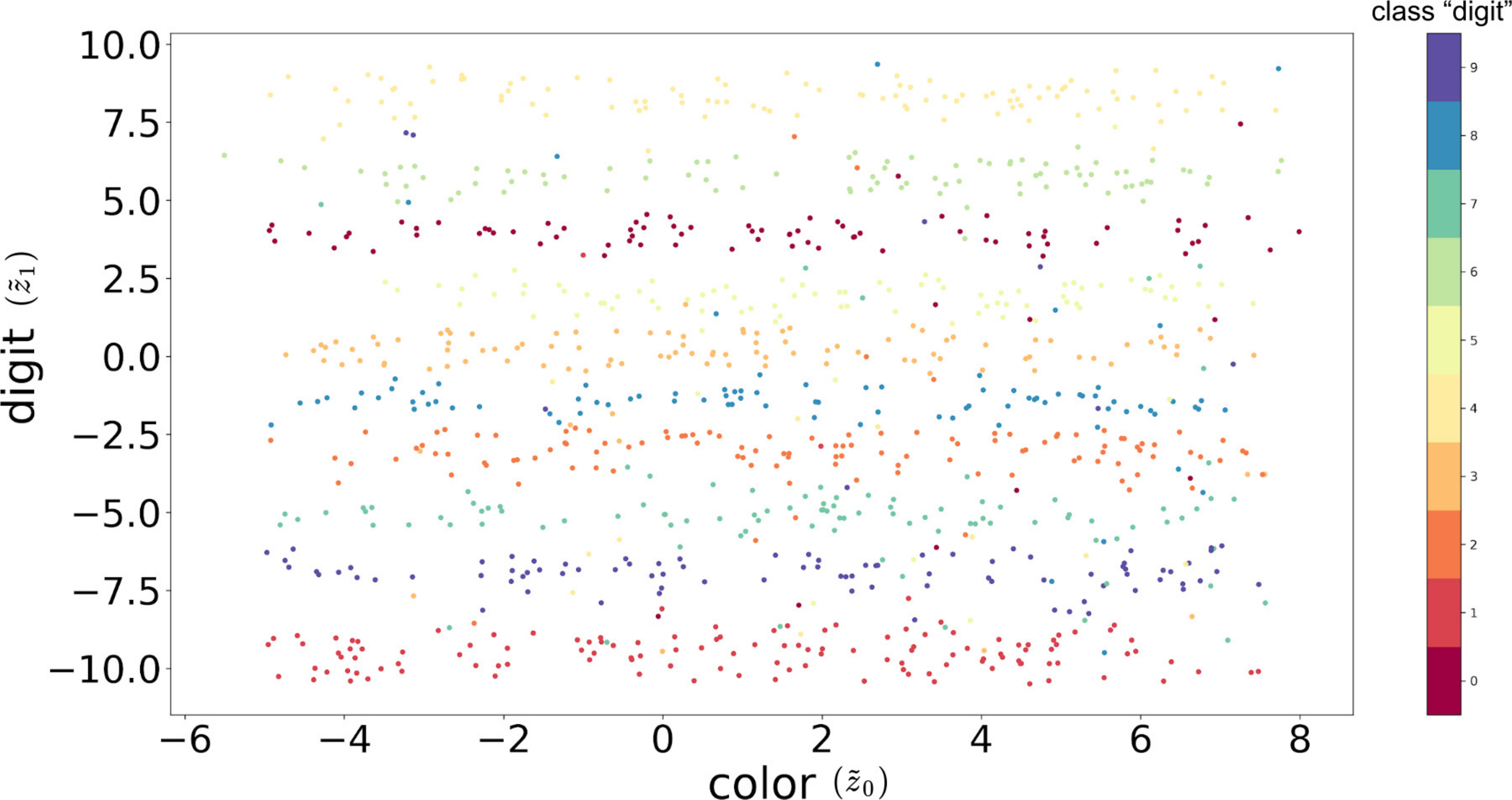}
  \caption{\emph{Left:} Output variance \textsl{per class} of a digit classifier on
  ColorMNIST, assessed via distribution of log-softmaxed logits and class
  predictions. $T$ disentangles $\tz_0$ (residual), $\tz_1$
  (digit) and $\tz_2$ (color). \emph{Right:} 1d disentangled UMAP embeddings of $\tz_1$ and $\tz_2$. See Sec.~\ref{sec:expclsf}.}
  \label{fig:cmnistresponse}
\end{figure*}
}

\newcommand{\resnetresponse}{
\begin{figure*}[!htb]
\centering
    \includegraphics[width=0.42\textwidth]{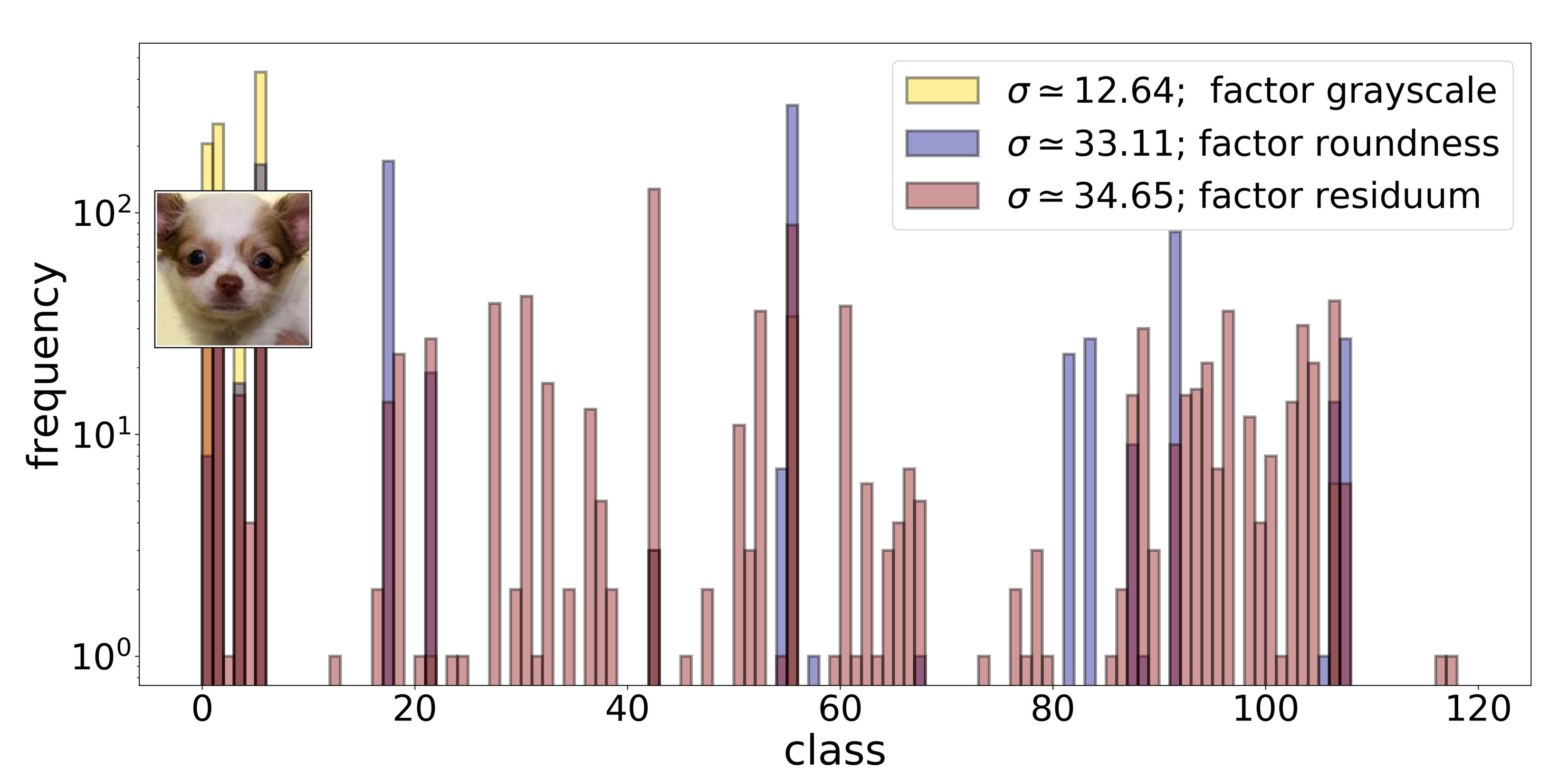}
    \includegraphics[width=0.42\textwidth]{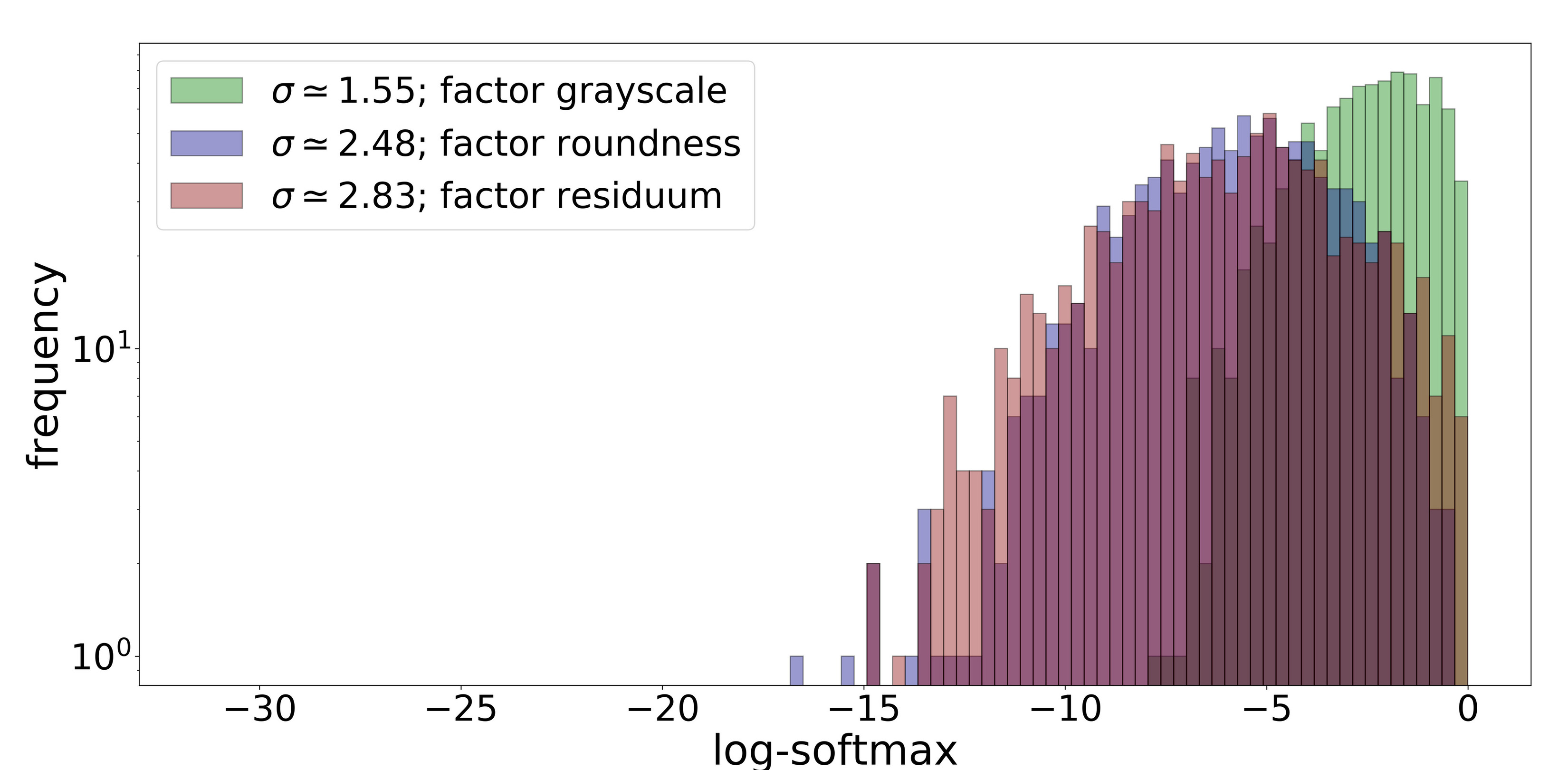}
  \caption{Output variance \textsl{per class} of a ResNet-50 classifier trained on AnimalFaces class identity, assessed via distribution of log-softmaxed logits and class
  predictions. $T$ is trained to disentangle $\tz_0$ (residual), $\tz_1$
  (roundness) and $\tz_2$ (greyscale). See Sec.~\ref{sec:expclsf}.}
  \label{fig:resnetresponse}
\end{figure*}
}

\newcommand{\cmnisttransfer}{
\begin{figure}[t]
\centering
\begin{minipage}[b]{0.15\textwidth}
    \centering
    \footnotesize $\tilde{z}_1$=``digit''\par\medskip
    \includegraphics[width=\textwidth]{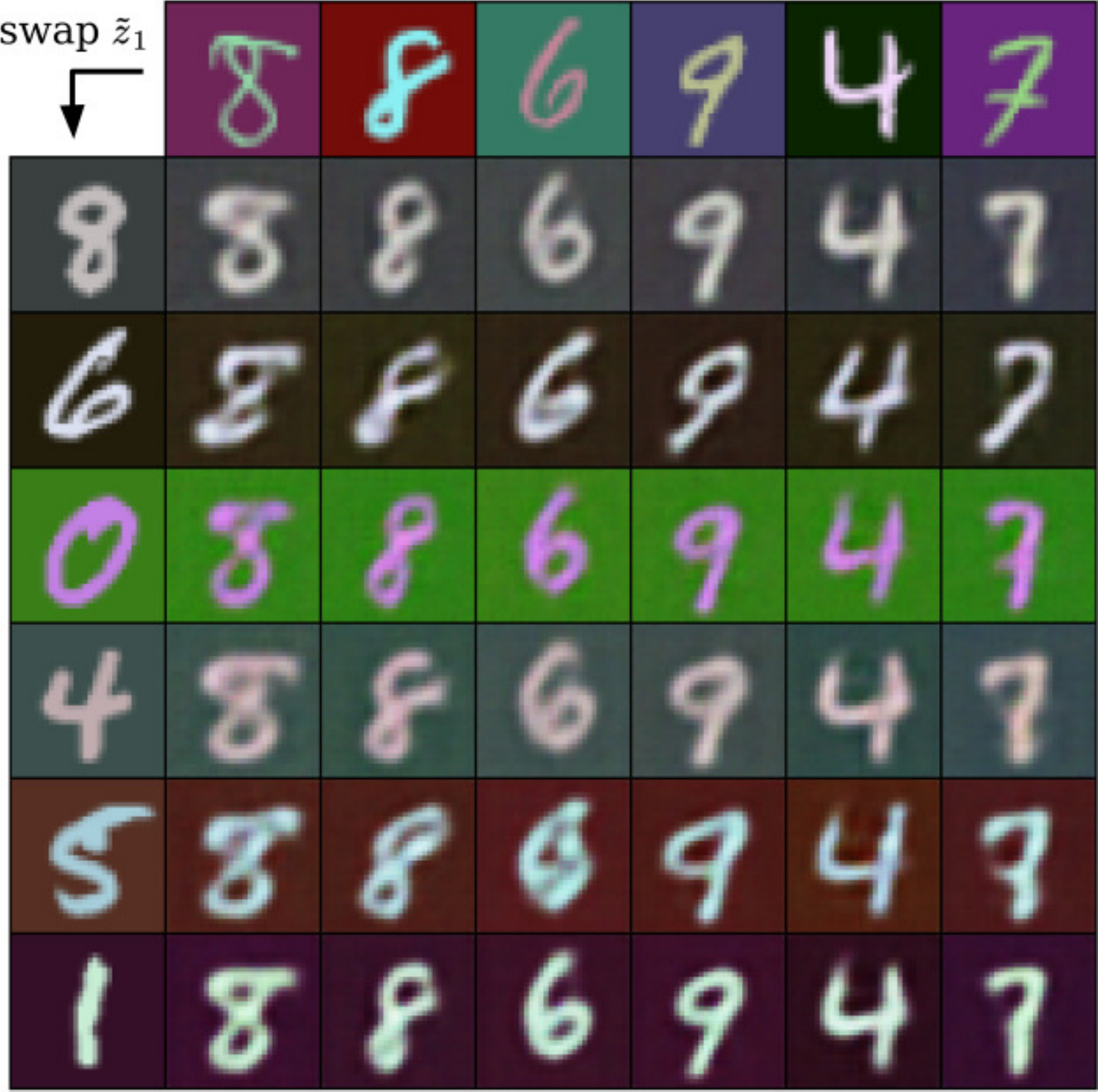}
  \end{minipage}
  \hfill
  \begin{minipage}[b]{0.15\textwidth}
    \centering
    \footnotesize$\tilde{z}_2$=``color''\par\medskip
    \includegraphics[width=\textwidth]{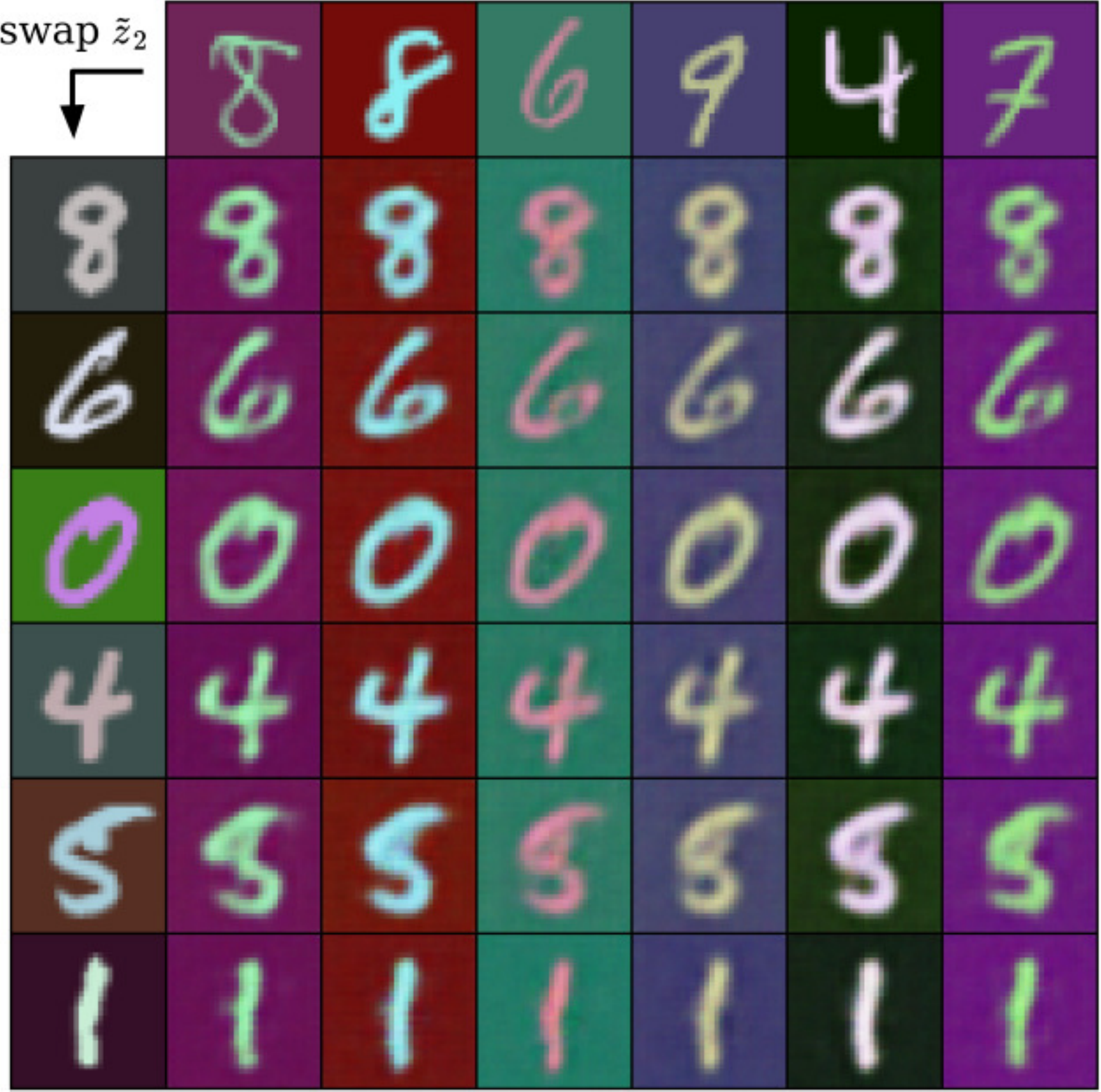}
  \end{minipage}
  \hfill
  \begin{minipage}[b]{0.15\textwidth}
    \centering
    \footnotesize $\tilde{z}_0$=``residual''\par\medskip
    \includegraphics[width=\textwidth]{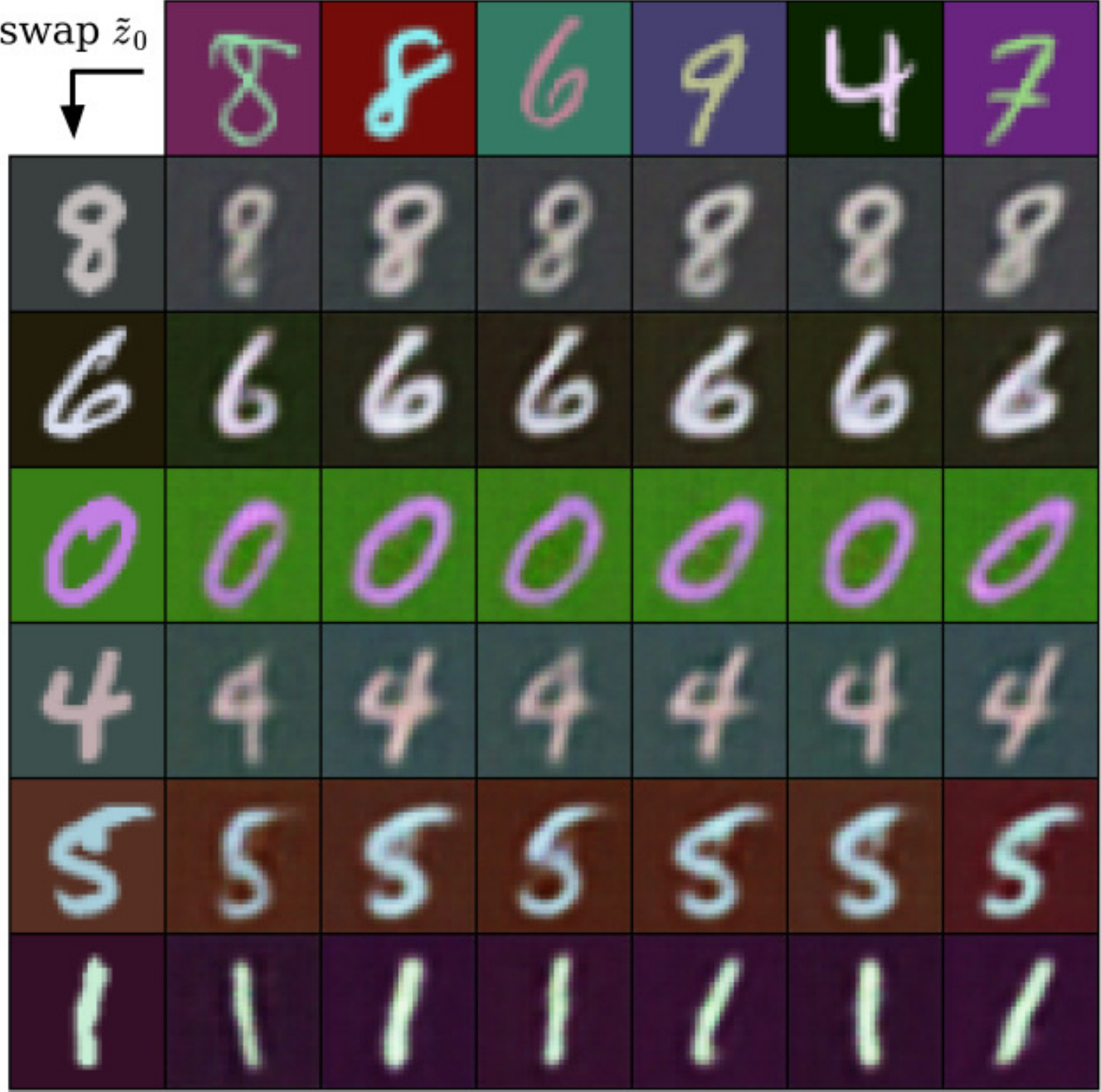}
  \end{minipage}
  \caption{
    Applied to latent representations $z$ of an autoencoder, our approach
    enables semantic image analogies. After transforming $z$ to disentangled
    semantic factors $(\tz_k)_{k=0}^{K} = T(z)$, we replace $\tz_k$ of the
    target image (leftmost column), with $\tz_k$ of the source image (top row).
    From left to right: $k=1$ (digit), $k=2$ (color), $k=0$ (residual).
  }
  \label{fig:cmnistswapping}
\end{figure}
}

\newcommand{\mnistinterpolationscompare}{
\begin{figure}[t]
\centering
\begin{minipage}[b]{0.47\textwidth}
    \includegraphics[width=\textwidth]{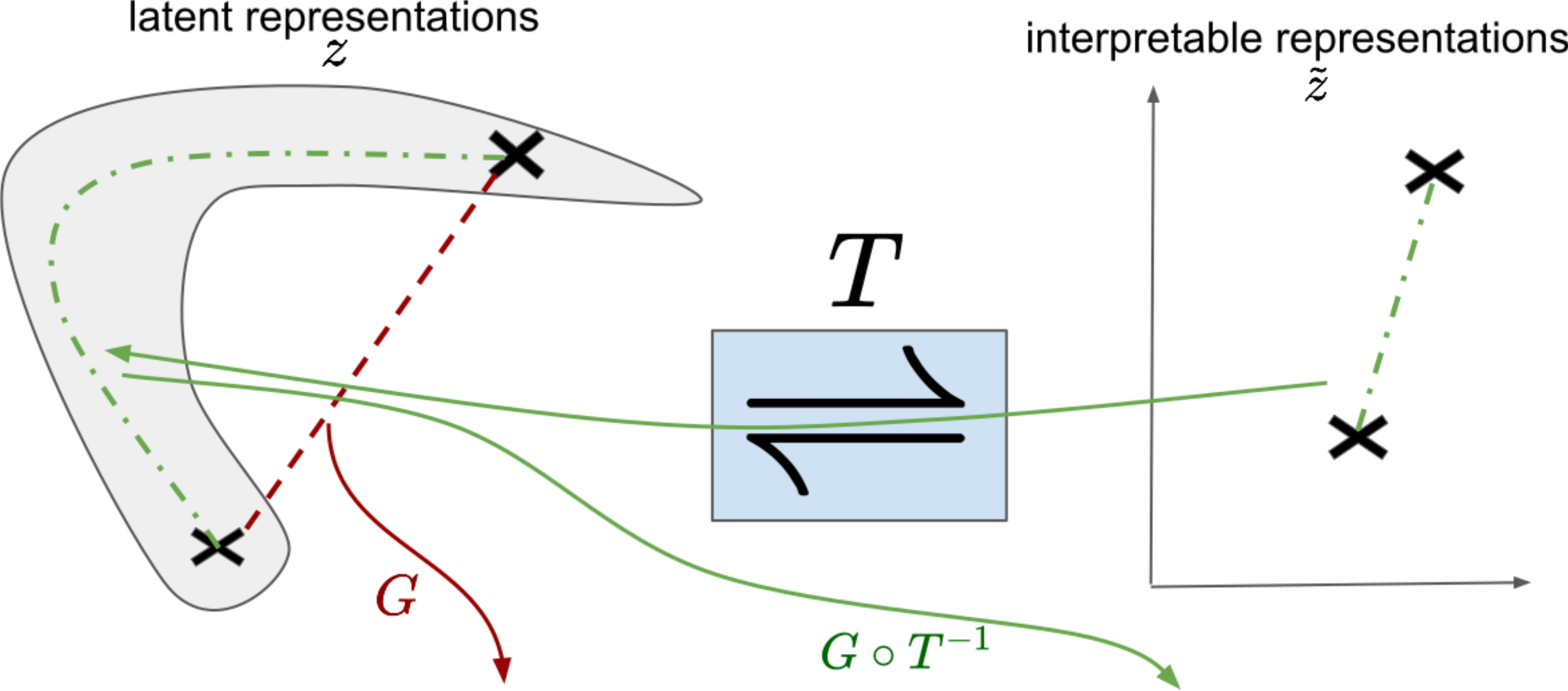}
  \end{minipage}
  \hfill
\begin{minipage}[b]{0.17\textwidth}
    \includegraphics[width=\textwidth]{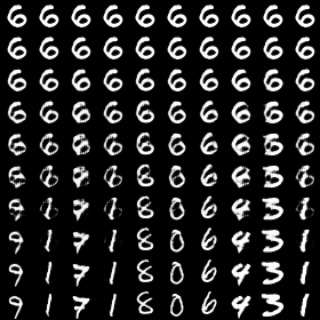}
    \centering
  \end{minipage}
  \hfill
  \begin{minipage}[b]{0.17\textwidth}
    \includegraphics[width=\textwidth]{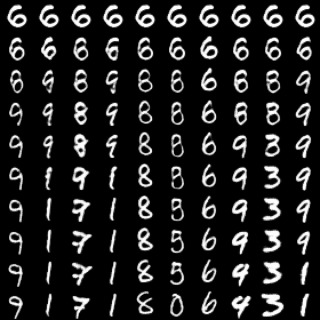}
    \centering
  \end{minipage}
  \caption{The inverse of our interpretation network $T$ maps linear walks
  in the interpretable domain back to nonlinear walks on the data manifold in
  the encoder space, which get decoded to meaningful images (bottom right). In contrast,
  decoded images of linear walks in the encoder space contain ghosting
  artifacts (bottom left).}
  \label{fig:mnistinterpolation}
\end{figure}
}
\newcommand{\animalswapping}{
\begin{figure}[t]
\centering
\includegraphics[width=0.345\textwidth]{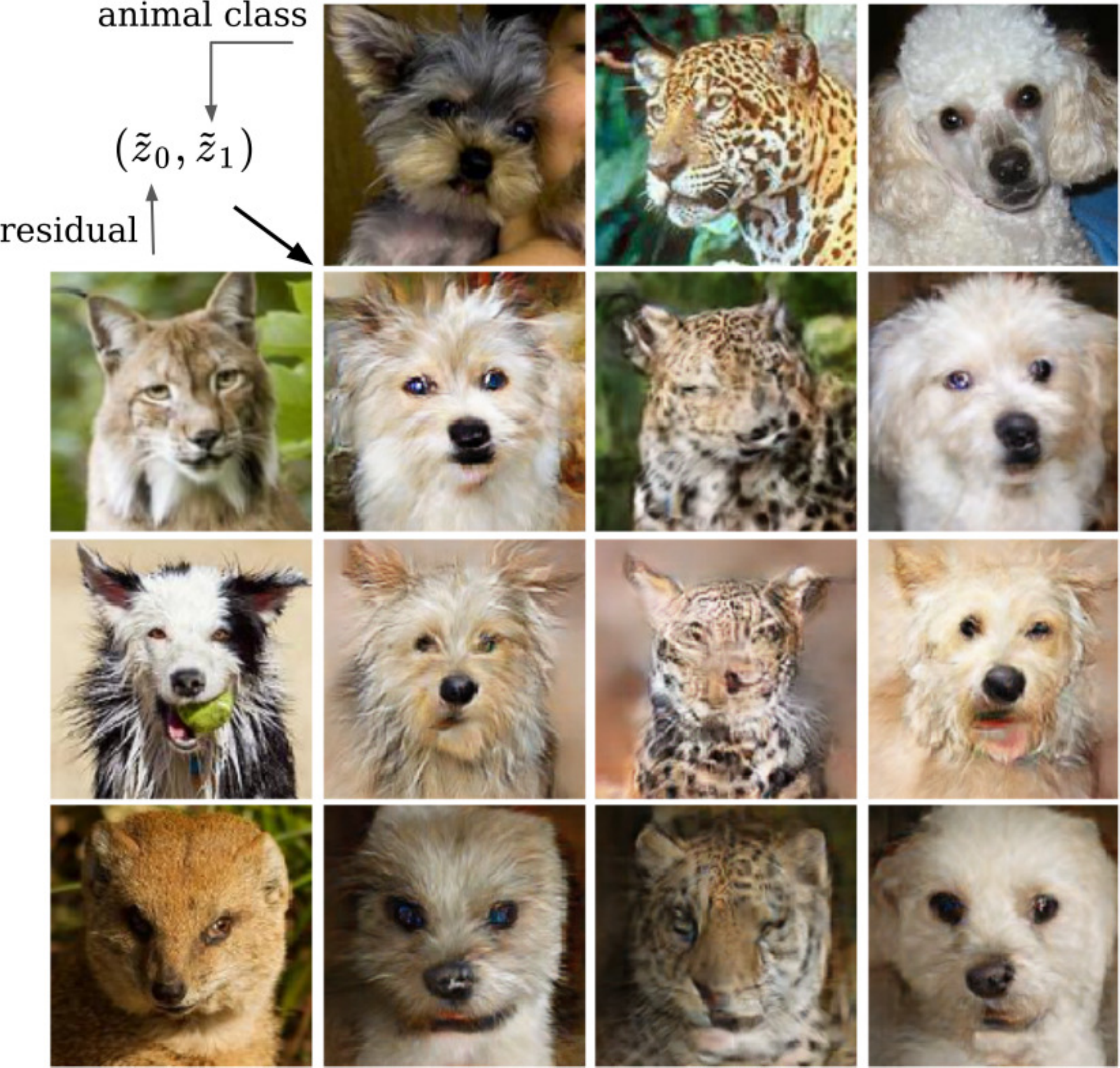}
  \caption{Transfer on AnimalFaces: We combine $\tz_0$ (residual) of the
  target image (leftmost column) with $\tz_1$ (animal class) of the source
  image (top row), resulting in a transfer of animal type from source to
  target.}
\label{fig:animalswapping}
\end{figure}
}
\newcommand{\dfswapping}{
\begin{figure}[t]
\centering
\includegraphics[width=0.345\textwidth]{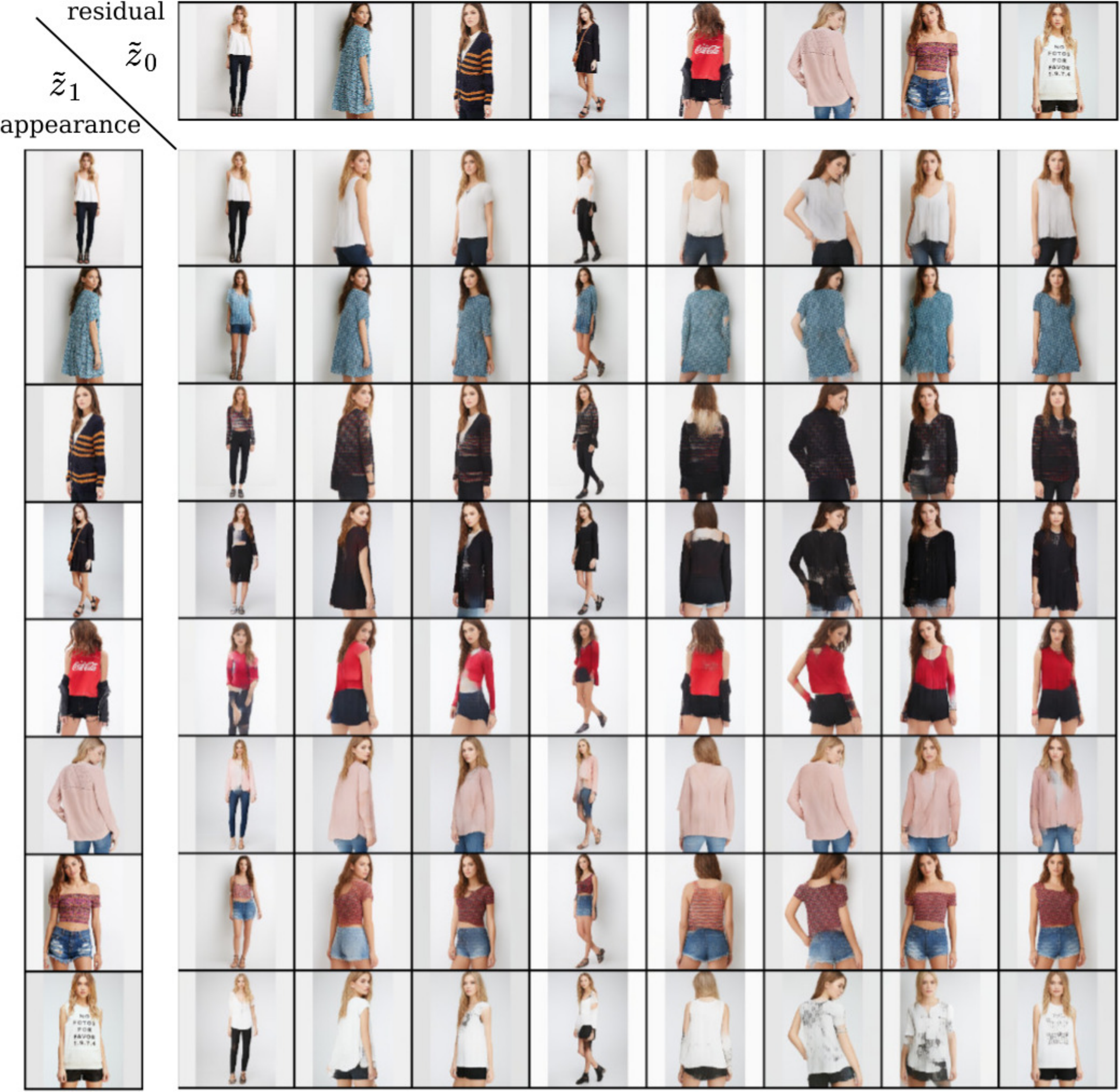}
  \caption{Transfer on DeepFashion: We combine $\tz_0$ (residual) of the
  target image (top row) with $\tz_1$ (appearance) of the source image
  (leftmost column), resulting in a transfer of appearances from source to
  target.}
  \label{fig:dfswapping}
\end{figure}
}
\newcommand{\dfappsampling}{
\begin{figure}[!htb]
\centering
\includegraphics[width=0.475\textwidth]{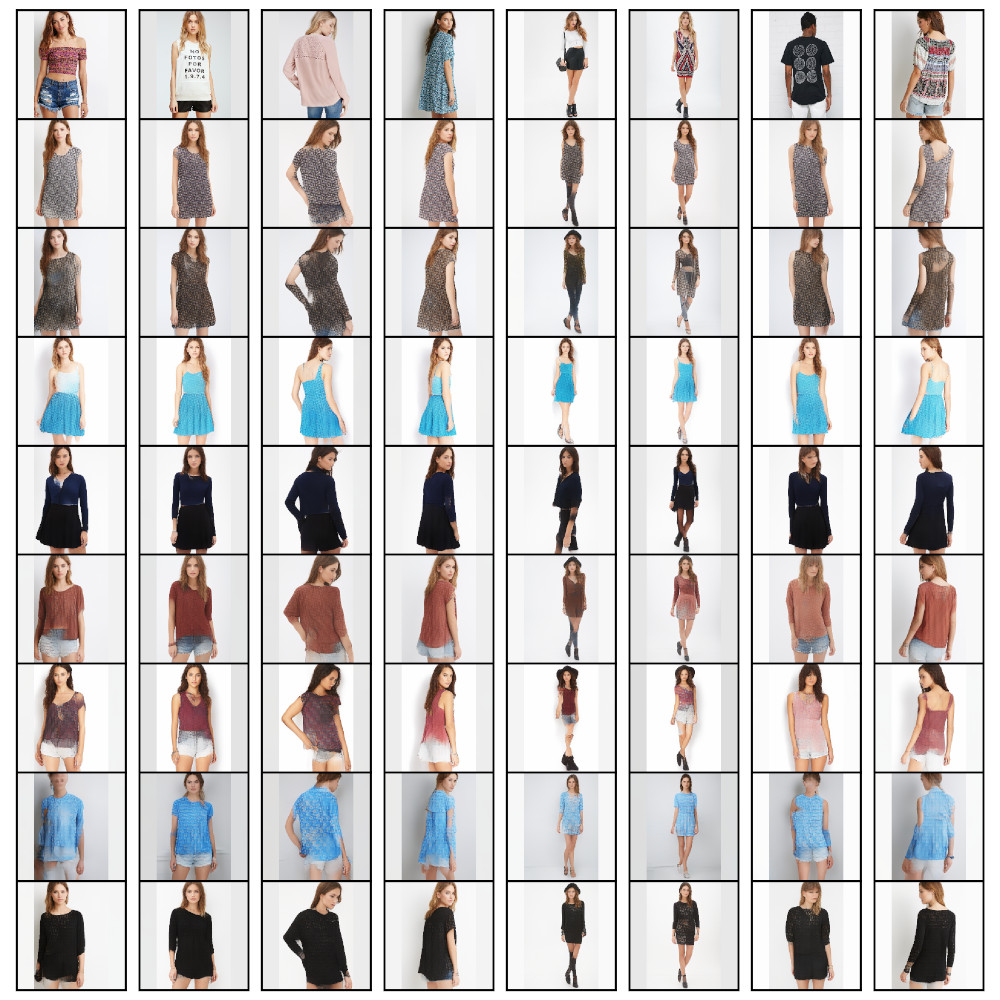}
\caption{Sampling on DeepFashion: Instead of swapping factors between
  images, we use the ability of our interpretation network to explore a
  factor's variability by directly sampling it. In each row, we combine a randomly
  sampled appearance factor $\tz_1$ with the residual factor $\tz_0$ of the
  topmost image.}
  \label{fig:dfsampling}
\end{figure}
}
\newcommand{\brownvsornstein}{
\begin{figure}[!htb]
\centering
\includegraphics[width=0.475\textwidth]{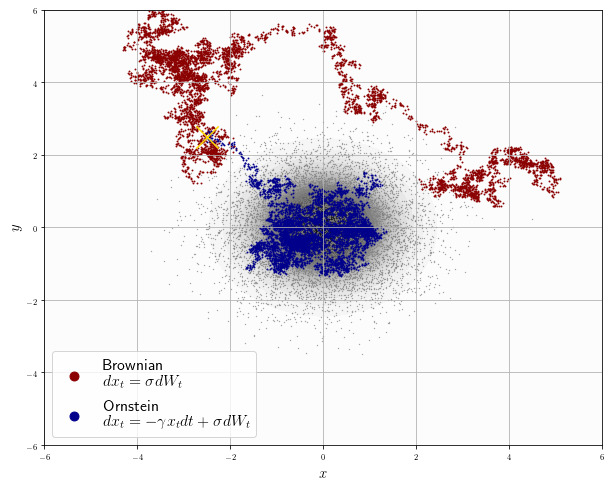}
\caption{Brownian motion in flat space (red) and in a harmonic potential
  (blue). The latter is known as \textsl{Ornstein-Uhlenbeck process} and
  used to explore a factor in proximity to a given reference point.}
\label{fig:brownornstein}
\end{figure}
}
\newcommand{\cmnistvid}{
\begin{figure}[!htb]
\centering
\includegraphics[width=0.475\textwidth]{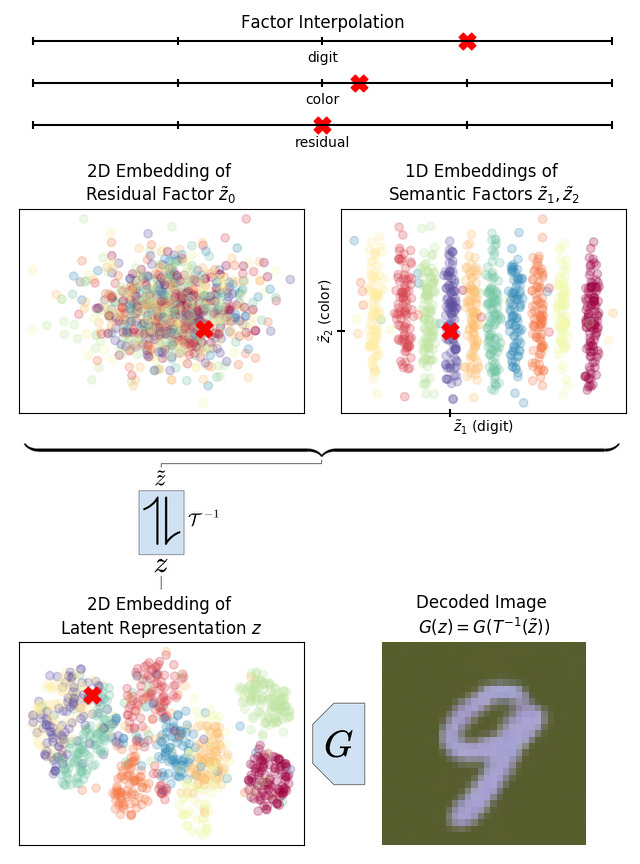}
\caption{Semantic image modifications and embeddings: We interpolate
  within individual semantic concepts and visualize representations embedded
  onto semantically-meaningful dimensions. See Sec.~\ref{sec:addres} for
  details and
  \href{https://compvis.github.io/iin/}{https://compvis.github.io/iin/} for an
  animated version.}
\label{fig:cmnistvid}
\end{figure}
}
\newcommand{\celebavid}{
\begin{figure}[!htb]
\centering
\includegraphics[width=0.475\textwidth]{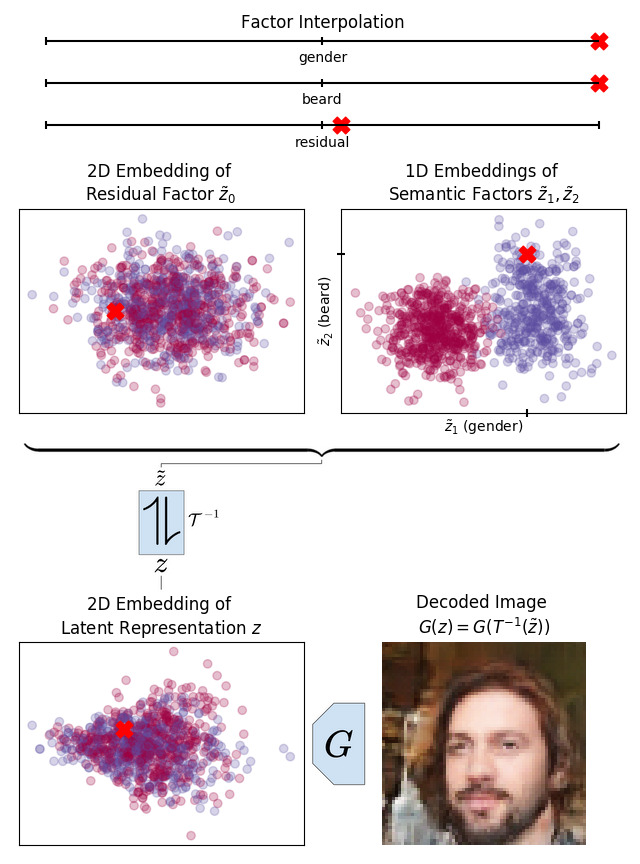}
\caption{Semantic image modifications and embeddings: We interpolate
  within individual semantic concepts and visualize representations embedded
  onto semantically-meaningful dimensions. See Sec.~\ref{sec:addres} for
  details and
  \href{https://compvis.github.io/iin/}{https://compvis.github.io/iin/} for an
  animated version.}
\label{fig:celebavid}
\end{figure}
}
\newcommand{\umapcmnist}{
\begin{figure}[b]
\centering
\begin{minipage}[b]{0.225\textwidth}
    \includegraphics[width=\textwidth]{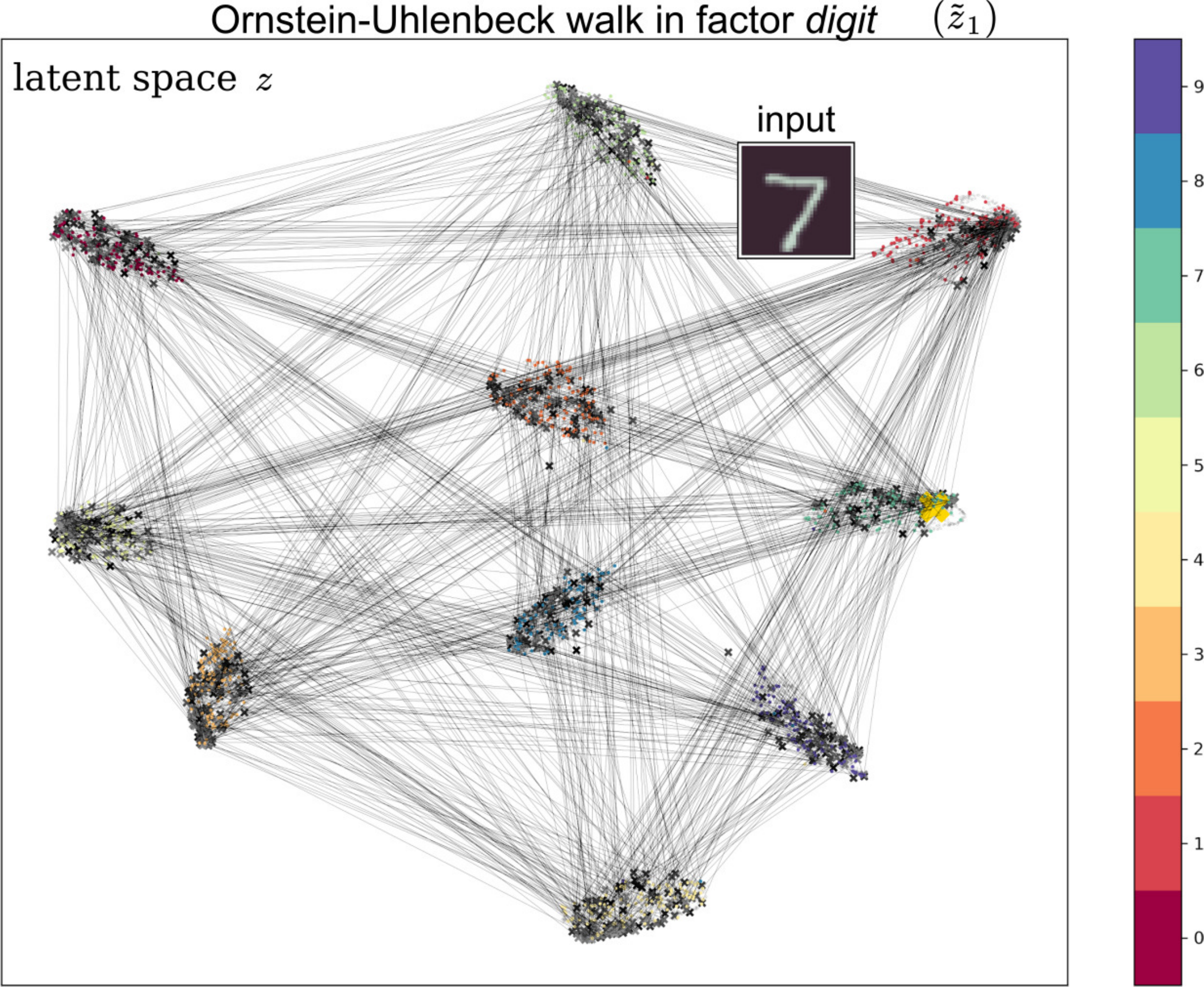}
  \end{minipage}
  \hfill
  \begin{minipage}[b]{0.225\textwidth}
          \includegraphics[width=\textwidth]{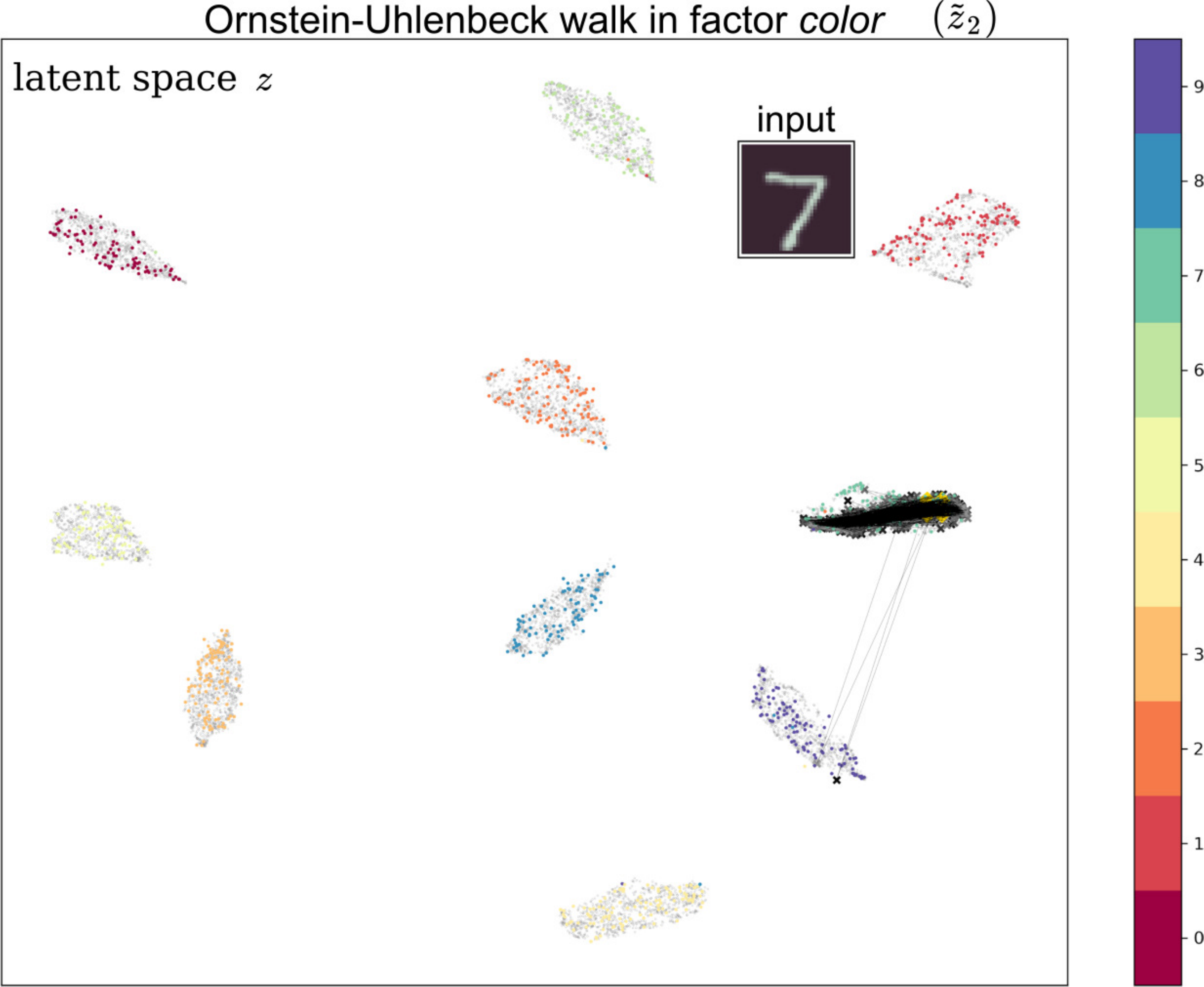}
  \end{minipage}
  \caption{
    UMAP embedding of a ColorMNIST classifier's latent space $z=E(x)$. 
    Colors of dots
    represent classes of test examples.
    We map latent representations $z$ to interpretable representations
    $\tz=T(z)$, where we perform a
    random walk
    in one of the factors
    $\tz_k$. Using $T^{-1}$, this random walk is mapped back to the latent space
    and shown as black crosses connected by gray lines.
    On the left, a random walk in the digit factor jumps between digit
    clusters, whereas on the right, a random walk in the color factor stays
    (mostly) within the digit cluster it starts from.
  }
  \label{fig:umapcmnist}
\end{figure}
}

\newcommand{\umapornsteincmnist}{
\begin{figure*}[!htb]
\centering
\begin{minipage}[b]{0.475\textwidth}
  \includegraphics[width=\textwidth]{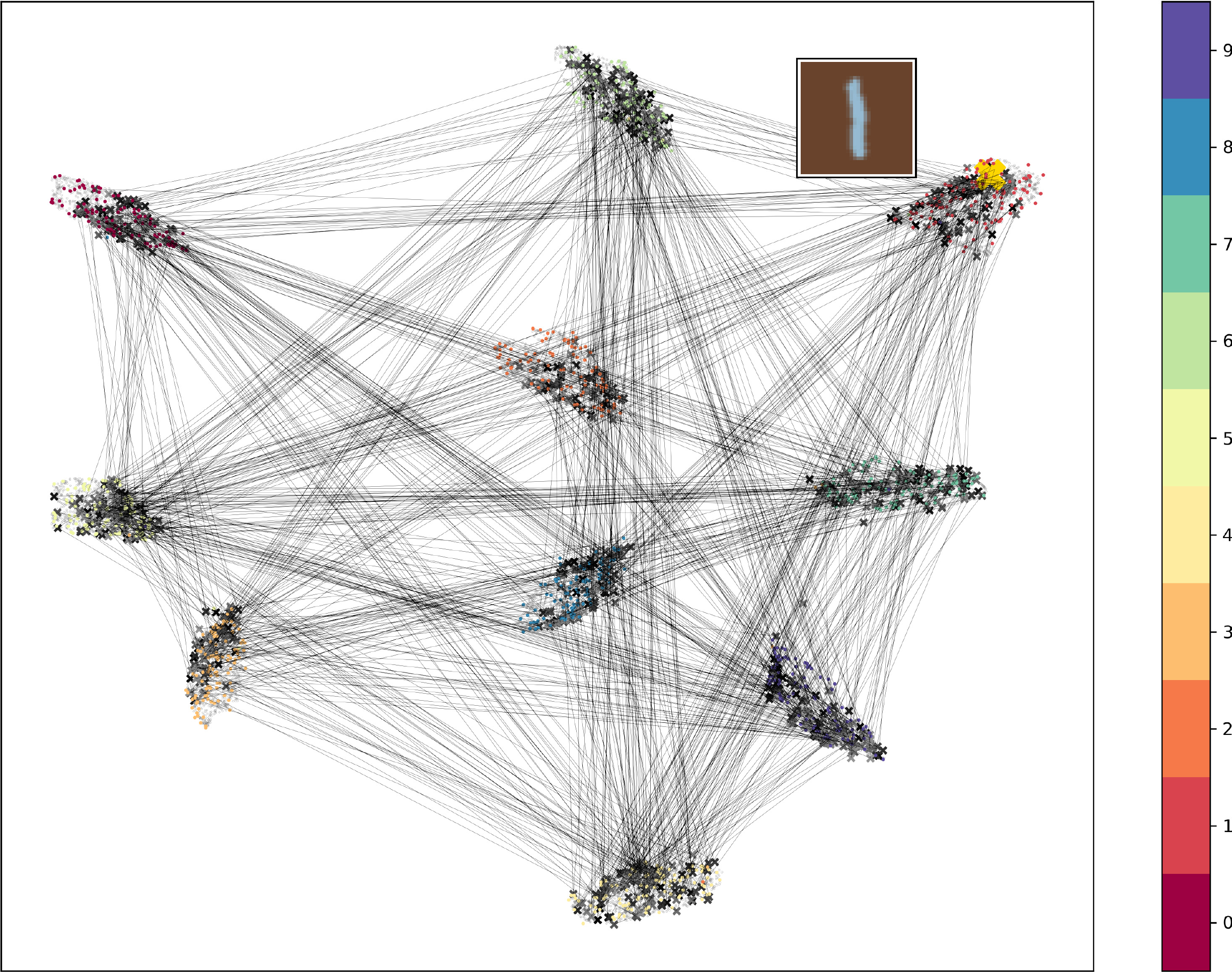}
  \end{minipage}
  \hfill
  \begin{minipage}[b]{0.475\textwidth}
          \includegraphics[width=\textwidth]{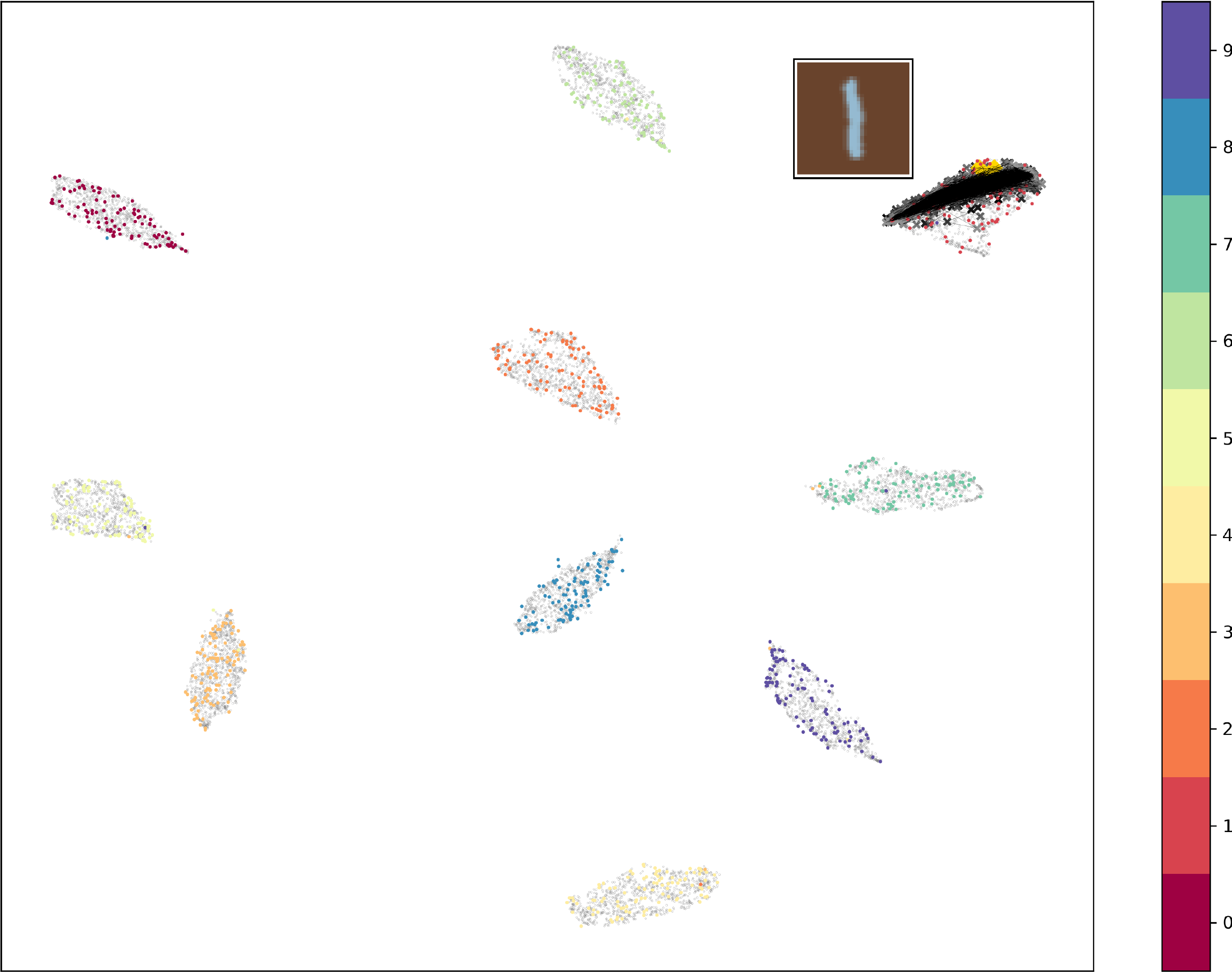}
  \end{minipage}
  \caption{Same as Fig.~\ref{fig:umapcmnist} for a different input example. Left: Changing factor \textsl{digit} in ColorMNIST classifier's hidden
  representation by a random walk in a harmonic potential. Right: Changing factor \textsl{color} in ColorMNIST classifier's hidden representation by a walk in a harmonic potential.}
\label{fig:umapornsteincmnist}
\end{figure*}
}

\newcommand{\umapresnet}{
\begin{figure*}[!htb]
\centering
\begin{minipage}[b]{0.33\textwidth}
    \includegraphics[width=\textwidth]{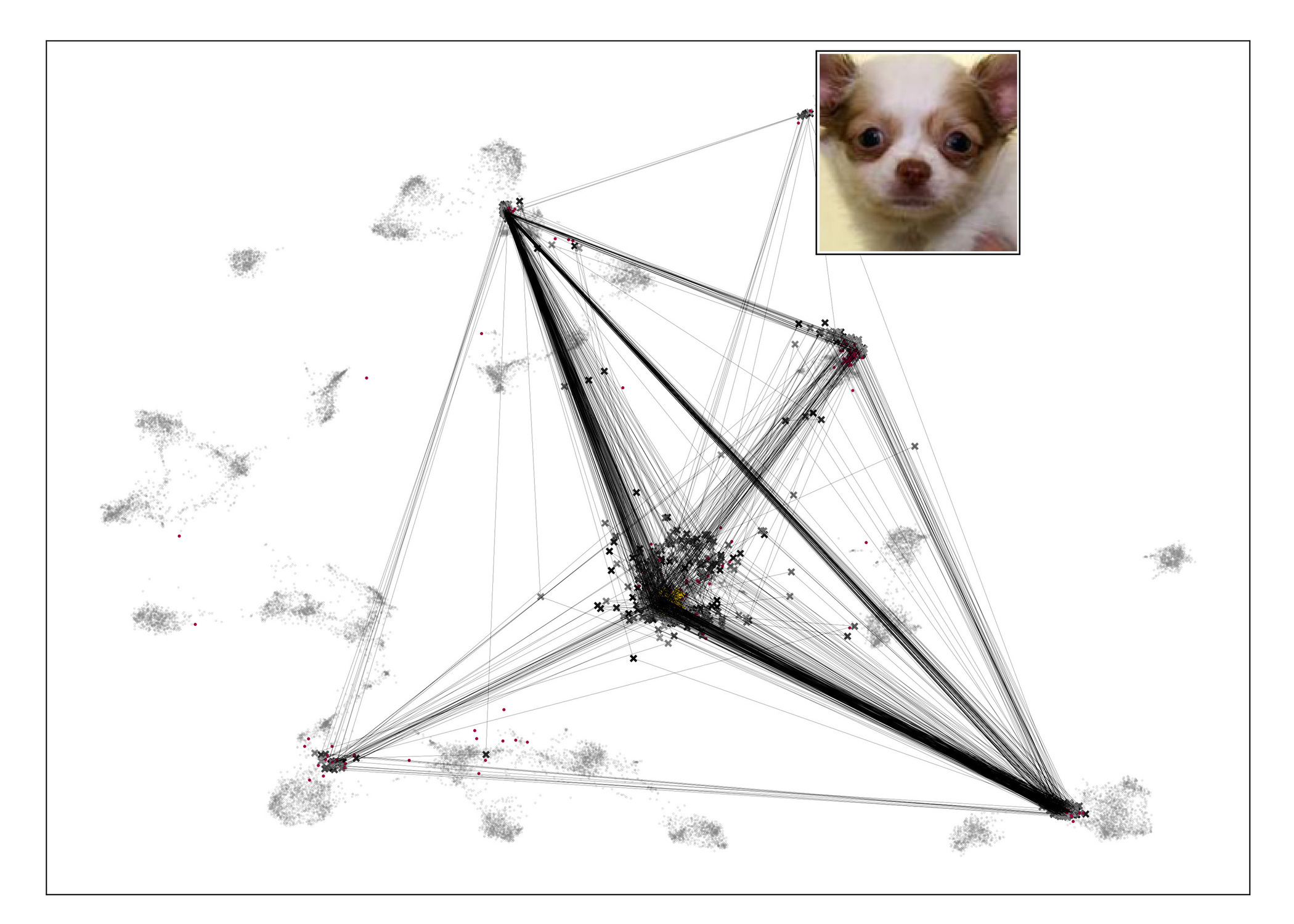}
\end{minipage}
  \hfill
\begin{minipage}[b]{0.33\textwidth}
    \includegraphics[width=\textwidth]{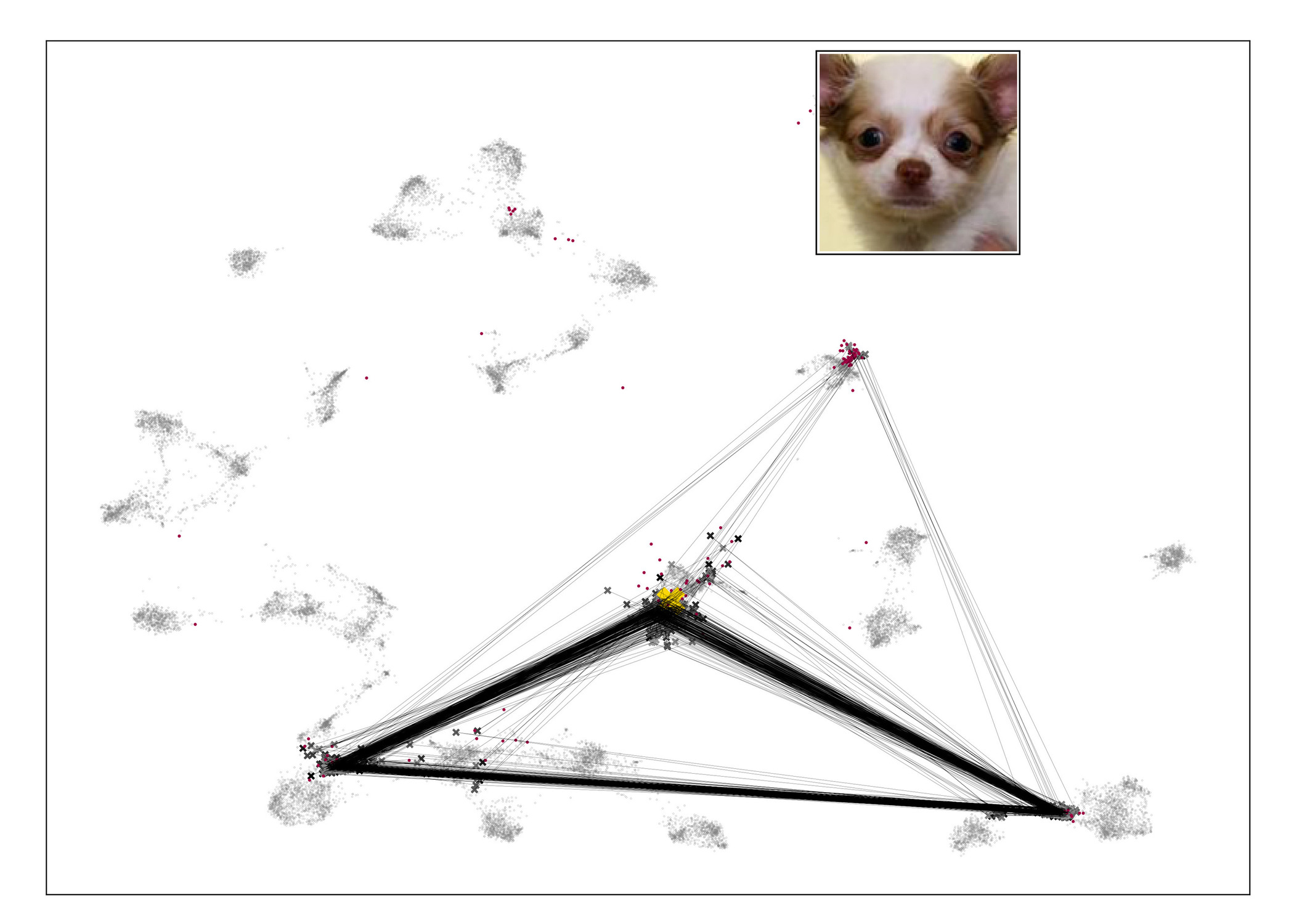}
\end{minipage}
\begin{minipage}[b]{0.33\textwidth}
    \includegraphics[width=\textwidth]{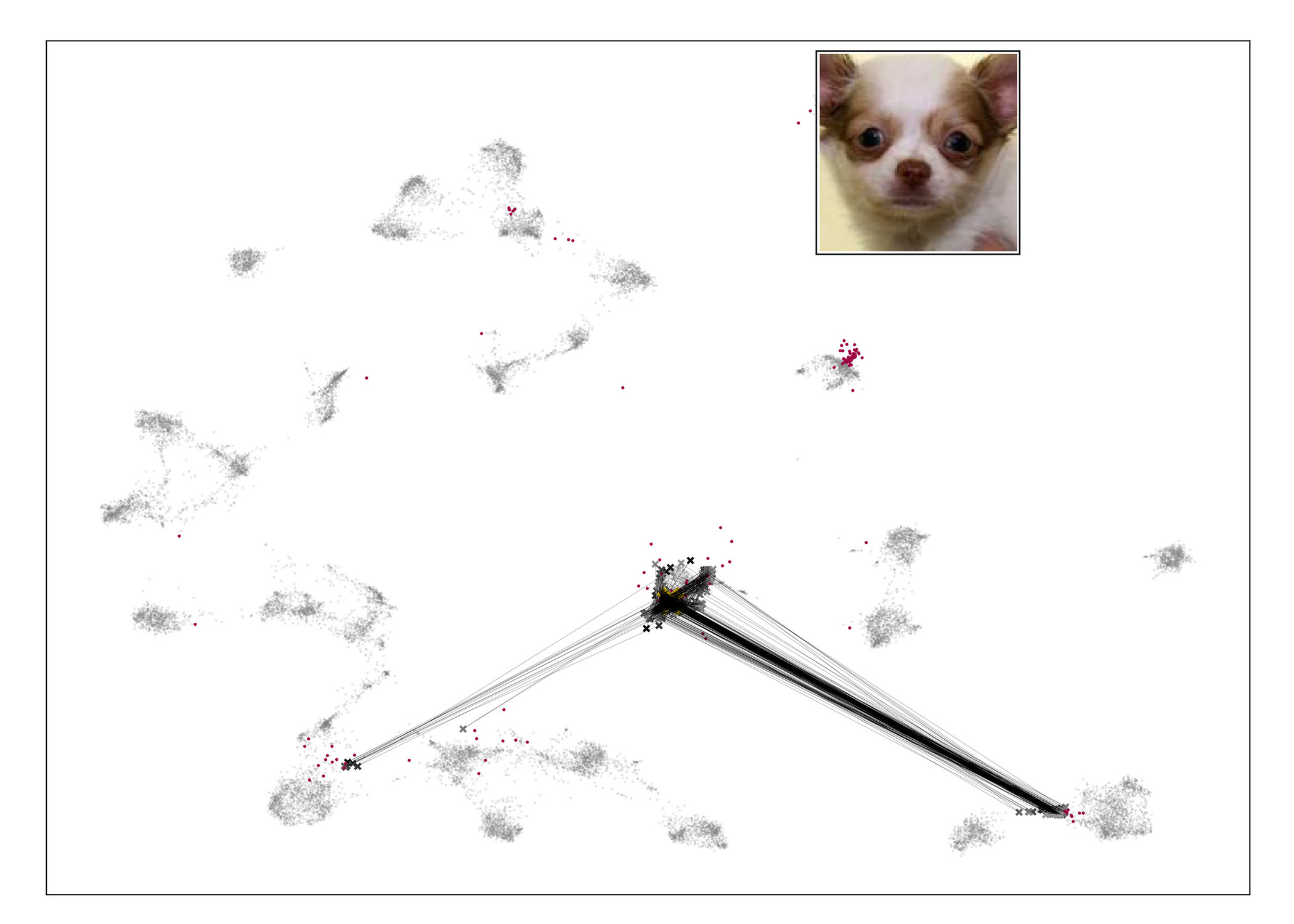}
\end{minipage}
\caption{Left: Changing factor \textsl{residual} in AnimalFaces classifier's hidden
  representation. Middle: Changing factor \textsl{roundness}. Right:
  Changing factor \textsl{grayness}.}
\label{fig:umapresnet}
\end{figure*}
}

\newcommand{\celebafid}{
\begin{table*}[t]
  \centering
\begin{tabular}{c|cccc}
\hline
   & MNIST & FashionMNIST & CIFAR-10 & CelebA \\
\hline
TwoStageVAE & $12.6\pm 1.5$  & $29.3 \pm 1.0$ &  $72.9\pm 0.9$ & $44.4 \pm 0.7 $ \\
WGAN GP  & $20.3 \pm 5.0$  & $24.5\pm 2.1$ & $55.8\pm0.9$ & $30.3 \pm 1.0$ \\
WGAN  & $6.7 \pm 0.4$  & $21.5\pm 1.6$ & $55.2\pm2.3$ & $41.3 \pm 2.0$ \\
DRAGAN  & $7.6 \pm 0.4$  & $27.7\pm 1.2$ & $69.8\pm2.0$ & $42.3 \pm 3.0$ \\
BEGAN  & $13.1 \pm 1.0$  & $22.9\pm 0.9$ & $71.9\pm1.6$ & $38.9 \pm 0.9$ \\
\hline
\hline
  \textbf{Ours} & $\mathbf{6.4\pm0.1}$ & $\mathbf{16.0\pm 0.1}$ & $\mathbf{45.7\pm0.3}$  & $\mathbf{20.2 \pm 0.5} $ \\
\hline 
\end{tabular}
\caption{FID scores of various AE-based and GAN models as reported in \citep{dai2019diagnosing}.}
\label{tab:fidbasic}
\end{table*}
}
\newcommand{\estdim}{
\begin{table}[b]
        \centering
  \begin{tabular}{ll||l|lll}
\hline
    Dataset & Model & Latent $z$ & \multicolumn{2}{l}{Interpretable $\tz$} \\
   &  & Dim. & Dim. & Factor $\tz_F$\\
\hline
  Color- & AE & 64 & 12 & Digit \\
  MNIST &  &  & 19 & Color \\
    \cline{2-5}
   & Classifier & 64 & 22 & Digit \\
   &  &  & 11 & Color \\
   \hline
\end{tabular}
\caption{Estimated dimensionalities of interpretable factors $\tz_F$
  representing different semantic concepts. Remaining dimensions are assigned
  to the residual
  factor $\tz_0$.
  Compared to an autoencoder, the color factor is smaller in
  case of a color-invariant classifier.
  }
\label{tab:estdim}
\end{table}
}

\newcommand{\numparams}{
\begin{table*}[!htb]
	 \centering
\begin{tabular}{|c|ccc||ccc|r|}
\hline
& \multicolumn{6}{c|}{Number of parameters per model $[10^6]$} & \\
\cline{2-7}
& \multicolumn{3}{c||}{ours} & \multicolumn{3}{c|}{TwoStageVAE \citep{dai2019diagnosing}} & \\
\cline{1-7}
input size & encoder & decoder & total & encoder & decoder & total & ratio\\
\hline
$32 \times 32$, $z \in \mathbb{R}^{64}$ & 3.0 & 2.9 & 5.9 & 8.7 & 8.6 & 17.3 & 0.34 \\
$64 \times 64$, $z \in \mathbb{R}^{64}$ & 3.8 & 3.3 & 7.1 & 33.9 & 33.8 & 67.7 & 0.10\\
$128 \times 128$, $z \in \mathbb{R}^{256}$ & 19.5 & 11.2 & 30.7 & (135.1) & (134.9) & (270.0) & 0.11 \\
\hline 
\end{tabular}
\caption{Comparison of the total number of parameters (in millions) used in the described autoencoding architectures. Note that the numbers in the last row for \citep{dai2019diagnosing} are calculated based on their architecture, but never used in their experiments.}
\label{tab:numparams}
\end{table*}
}

\newcommand{\hyperparamsiin}{
\begin{table}[!htb]
\centering
\begin{tabular}{|c|cccc|}
\hline
input size & $n_{flow}$ & $H$ & $D$ & $\sigma_{ab}$ \\
\hline
$z \in \mathbb{R}^{64}$ & 12 & 512 & 2 & 0.9\\
$z \in \mathbb{R}^{256}$ & 12 & 512 & 2 & 0.9\\
$z \in \mathbb{R}^{2048}$ & 6 & 2048 & 2 & 0.9\\
\hline 
\end{tabular}
\caption{Hyperparameters used for training the transformer $T$. See Sec.~\ref{sec:hyperflow} for a definition of the notation used.}
\label{tab:hyperparamsiin}
\end{table}
}

\newcommand{\aearchitecture}{
\begin{table*}[!htb]
	 \centering
\begin{tabular}{|l|l|}
\hline
$E$: \textbf{Encoder} & $G$: \textbf{Decoder} \\
\hline
Conv2d(3, 64, 4, 2, 1), ActNorm, LeakyReLU(0.2) & ConvT2d($z$, 512, $h/16$, 1, 0), ActNorm, LeakyReLU(0.2)\\
Conv2d(64, 128, 4, 2, 1), ActNorm, LeakyReLU(0.2) & ConvT2d(512, 256, 4, 2, 1), ActNorm, LeakyReLU(0.2) \\
Conv2d(128, 256, 4, 2, 1), ActNorm, LeakyReLU(0.2) &  ConvT2d(256, 128, 4, 2, 1), ActNorm, LeakyReLU(0.2) \\
Conv2d(256, 512, 4, 2, 1), ActNorm, LeakyReLU(0.2) &  ConvT2d(128, 64, 4, 2, 1), ActNorm, LeakyReLU(0.2) \\
Conv2d(512, $2\cdot z$, $h/16$, 1, 0) & ConvT2d(64, 3, 4, 2, 1), Tanh \\
\hline 
\end{tabular}
\caption{Architecture of our autoencoder-model. We assume quadratic images,
  \ie $h=w$ for an image of size $c \times h \times w$ with $c$ channels.}
\label{tab:archae}
\end{table*}
}

\newcommand{\clfarchitecture}{
\begin{table*}[!htb]
	 \centering
\begin{tabular}{|l|l|}
\hline
$E$: \textbf{Encoder} & $G$: \textbf{Classification Head} \\
\hline
Conv2d(3, 64, 4, 2, 1), ActNorm, LeakyReLU(0.2) & Linear($z$, $n_{classes}$)\\
Conv2d(64, 128, 4, 2, 1), ActNorm, LeakyReLU(0.2) & \\
Conv2d(128, 256, 4, 2, 1), ActNorm, LeakyReLU(0.2) &   \\
Conv2d(256, 512, 4, 2, 1), ActNorm, LeakyReLU(0.2) &  \\
Conv2d(512, $z$, $h/16$, 1, 0) & \\
\hline 
\end{tabular}
\caption{Architecture of our basic classification model.}
\label{tab:archclf}
\end{table*}
}

\begin{document}

\title{A Disentangling Invertible Interpretation Network for \\
Explaining Latent Representations}
\author{Patrick Esser\thanks{Both authors contributed equally to this
work.} \qquad Robin Rombach\footnotemark[1] \qquad Bj\"orn Ommer\\
Heidelberg Collaboratory for Image Processing\\
IWR, Heidelberg University, Germany\\
}

\maketitle
\thispagestyle{empty}

\begin{abstract}
Neural networks have greatly boosted performance in computer vision by
learning powerful representations of input data. The drawback of
end-to-end training for maximal overall performance are black-box models
whose hidden representations are lacking interpretability:
Since distributed coding is optimal for latent layers to improve their
robustness, attributing meaning to parts of a hidden feature vector or to
individual neurons is hindered.
We formulate interpretation as a translation of hidden representations onto
semantic concepts that are comprehensible to the user. The mapping between
both domains has to be bijective so that semantic modifications in the
target domain correctly alter the original representation. The proposed
invertible interpretation network can be transparently applied on top of
existing architectures with no need to modify or retrain them.
Consequently, we translate an original representation to an equivalent yet
interpretable one and backwards without affecting the expressiveness and
performance of the original. The invertible interpretation network
disentangles the hidden representation into separate, semantically
meaningful concepts. Moreover, we present an efficient approach to define
semantic concepts by only sketching two images and also an unsupervised
strategy. Experimental evaluation demonstrates the wide applicability to
interpretation of existing classification and image generation networks as
well as to semantically guided image manipulation.
\end{abstract}

\section{Introduction}
Deep neural networks have achieved unprecedented performance in various
computer vision tasks \cite{xie2017aggregated,Huang_2017} by learning
task-specific hidden representations rather
than relying on predefined hand-crafted image features. However, the
significant performance boost due to end-to-end learning comes at the cost
of now having black-box models that are lacking interpretability: A deep
network may have found a solution to a task, but human users cannot
understand the causes for the predictions that the model makes
\cite{Miller_2019}. Conversely, users must also be able to understand what
the hidden representations have \emph{not} learned and on which data the
overall model will, consequently, fail. Interpretability is therefore a
prerequisite for safeguarding AI, making its decisions transparent to the
users it affects, and understanding its applicability, limitations, and the
most promising options for its future improvement.
\overview %

\renewcommand{\topfraction}{0.5}

A key challenge is that learned latent representations typically do not
correspond to semantic concepts that are comprehensible to human users.
Hidden layer neurons are trained to help in solving an overall task in the
output layer of the network. Therefore, the output neurons correspond to
human-interpretable concepts such as object classes in semantic image
segmentation \cite{chen2014semantic} or bounding boxes in object detection
\cite{szegedy2013deep}. In contrast, the
hidden layer representation of semantic concepts is a distributed pattern
\cite{Fong_2018}. This distributed coding is crucial for the
robustness and generalization performance of neural networks despite noisy
inputs, large intra-class variability, and the stochastic nature of
the learning algorithm \cite{Hinton1986DistributedR}. However, as a downside of semantics being
distributed over numerous neurons it is impossible
to attribute semantic
meaning to only an individual neuron despite attempts to backpropagate
\cite{Montavon_2017} or
synthesize \cite{simonyan2013deep,yosinski2015understanding} their
associated semantic concepts. One solution has been to modify and constrain the
network so that abstract concepts can be localized in the hidden
representation \cite{Zhou_2016}. However, this alters the network architecture and
typically deteriorates overall performance \cite{Ribeiro_2016}.

\textbf{Objective:} Therefore, our goal needs to be an approach that can be
transparently applied on top of arbitrary existing networks and their
already learned representations without altering them. We seek a
\emph{translation} between these hidden representations and human-comprehensible semantic concepts---a non-linear mapping between the two
domains. This translation needs to be invertible, i.e. an invertible neural
network (INN) \cite{dinh2014nice,dinh2016density,jacobsen2018irevnet,kingma2018glow}, so that modifications in the domain of semantic concepts
correctly alter the original representation.

To interpret a representation, we need to attribute meaning to parts of the
feature encoding. That is, we have to disentangle the high-dimensional
feature vector into multiple multi-dimensional factors so that each is
mapped to a separate semantic concept that is comprehensible to the user. As
discussed above, this disentangled mapping should be bijective so that
modifications of the disentangled semantic factors correctly translate back
to the original representation. We can now, without any supervision,
disentangle the representation into independent concepts so that a user can
post-hoc identify their meaning. Moreover, we present an efficient strategy
for defining semantic concepts. It only requires two sketches that exhibit a
change in a concept of interest rather than large annotated training sets
for each concept. Given this input, we derive the invariance properties
that characterize a concept and generate synthesized training data to train
our invertible interpretation network. This network then acts as a
translator that disentangles the original representation into multiple
factors that correspond to the semantic concepts.

Besides interpreting a network representation, we can also interpret the
structure that is hidden in a dataset and explain it to the user. Applying the
original representation and then translating onto the disentangled semantic
factors allows seeing which concepts explain the data and its variability.
Finally, the invertible translation supports semantically meaningful
modifications of input images: Given an autoencoder representation, its
representation is mapped onto interpretable factors, these can be modified and
inverse translation allows to apply the decoder to project back into the image
domain. In contrast to existing disentangled image synthesis
\cite{ma2018disentangled, esser2018variational, lorenz2019unsupervised,
kotovenko2019content,esser2019unsupervised}, our invertible approach can be
applied on top of existing autoencoder representations, which therefore do not
have to be altered or retrained to handle different semantic concepts.
Moreover, for other architectures such as classification networks,
interpretability helps to analyze their invariance and robustness.

To summarize, \emph{(i)} we present a new approach to the interpretability of
neural networks, which can be applied to arbitrary existing models without
affecting their performance; \emph{(ii)} we obtain an invertible translation
from hidden representations to disentangled representations of semantic
concepts; \emph{(iii)} we propose a method that allows users to efficiently
define semantic concepts to be used for our interpretable representation;
\emph{(iv)} we investigate the interpretation of hidden representations, of
the original data, and demonstrate semantic image modifications enabled by
the invertibility of the translation network.

\section{Interpretability}

An interpretation is a translation between two domains such that concepts of
the first domain can be understood in terms of concepts of the second
domain. Here, we are interested in interpretations of internal
representations of a neural network in terms of human-understandable
representations. Examples for the latter are given by textual descriptions,
visual attributes or images.

To interpret neural networks, some approaches modify network
architectures or losses used for training to obtain inherently more
interpretable networks. \cite{Zhou_2016} relies on a global average pooling
layer to obtain class activation maps, \ie heatmaps showing which regions of
an input are most relevant for the prediction of a certain object class.
\cite{Zhang_2018} learn part specific convolutional filters by restricting
filter activations to localized regions. Invertible neural networks
\cite{dinh2014nice,dinh2016density,jacobsen2018irevnet,kingma2018glow} have
been used to get a better understanding of adversarial attacks
\cite{jacobsen2018excessive}. Instead of replacing existing architectures
with invertible ones, we propose to augment them with invertible
transformations. Using the invertibility, we can always map back and forth
between original representations and interpretable ones without loss of
information. Thus, our approach can be applied to arbitrary existing
architectures without affecting their performance, whereas approaches
modifying architectures always
involve a trade-off between interpretability and performance.

Most works on interpretability of existing networks focus on visualizations.
\cite{Zeiler_2014} reconstruct images which activated a specific feature
layer of a network. 
\cite{simonyan2013deep}
uses gradient ascent to synthesize images which maximize class probabilities
for different object classes.
\cite{yosinski2015understanding} generalizes
this to arbitrary
neurons within a network. Instead of directly optimizing over pixel values,
\cite{nguyen2016synthesizing}
optimize over input codes of a generator network which was trained to
reconstruct images from hidden layers. 
\cite{Zhou_2016}
avoid synthesizing images from scratch and look for regions within
a given image that activate certain neurons. %
For a specific class of functions, \cite{bach2015pixel} decompose the
function into relevance scores which can be visualized pixel-wise.
Layer-wise relevance propagation \cite{Montavon_2017} is a more general
approach to propagate relevance scores through a network based on rules to
distribute the relevance among input neurons. \cite{petsiuk2018rise} shows
how saliency maps representing the importance of pixels for a classifier's
decision can be obtained without access to the classifiers gradients.
All these approaches assume that a fixed set of
neurons is given and should be interpreted in terms of inputs which activate
them. However, %
\cite{Bau_2017}, \cite{Fong_2018}
demonstrated that %
networks use distributed %
representations.
In particular, semantic concepts are encoded by activation patterns of
multiple neurons and single neurons are not concept specific but involved in
the representation of different concepts. We directly address this finding
by learning a non-linear transformation from a distributed representation to
an interpretable representation with concept specific factors.

While \cite{Fong_2018} shows that for general networks we must expect
internal representations to be distributed, there are situations where
representations can be expected to have a simpler structure: Generative
models are trained with the explicit goal to produce images from samples of
a simple distribution, \eg a Gaussian distribution. Most
approaches are based either on Variational Autoencoders \cite{VAE,VAE2},
which try to reconstruct images from a representation whose marginal
distribution is matched to a standard normal distribution, or on Generative
Adversarial Networks \cite{gan,hoang2018mgan,pan2018latent}, which directly map samples from a standard
normal distribution to realistic looking images as judged by a
discriminator network. The convexity of the Gaussian density makes linear
operations between representations meaningful. Linear interpolations between
representations enable walks along nonlinear data manifolds
\cite{radford2015unsupervised}. \cite{larsen2015autoencoding} finds
visual attribute vectors which can be used to interpolate between binary
attributes. To this end, two sets of images containing examples with or
without an attribute are encoded to their representations, and the direction
between their means is the visual attribute vector. Such attribute vectors
have also been found for classifier networks \cite{upchurch2017deep}, but
because their representations have no linear structure, the approach is
limited to aligned images. \cite{radford2015unsupervised,NIPS2015_5845}
demonstrated that vector arithmetic also enables analogy making.
\cite{shen2019interpreting} interprets the latent space of a GAN by finding
attribute vectors as the normal direction of the decision boundary of an
attribute classifier. \cite{goetschalckx2019ganalyze} uses a similar
approach to find attribute vectors associated with cognitive properties such
as memorability, aesthetics and emotional valence. While these approaches
provide enhanced interpretability through modifcation of attributes they are
limited to representations with a linear structure. In contrast, we provide
an approach to map arbitrary representations into a space of interpretable
representations. This space consists of factors representing semantic
attributes and admits linear operations. Thus, we can perform semantic
modifications in our interpretable space and, due to the invertibility of
our transformation, map the modified representation back to the original
space.

\section{Approach}
\newcommand{\Normal}{\mathcal{N}}
\newcommand{\Id}{\mathbf{1}}
\newcommand{\covar}{\sigma}
\newcommand{\covariance}[1]{\text{Cov}\bigl(#1\bigr)}
\newcommand{\variance}[1]{\text{Var}(#1)}
\newcommand{\expect}{\mathbb{E}}
\newcommand{\loss}{\mathcal{L}}
\newcommand{\lossunc}{\mathcal{L}_{\text{unsup}}}
\newcommand{\cond}{\vert}
\DeclarePairedDelimiter\abs{\lvert}{\rvert}%

\subsection{Interpreting Hidden Representations}
\label{sec:train}

\paragraph{Invertible Transformation of Hidden Representations:}
Let $f$ be a given neural network to be interpreted.  We place no restrictions
on the network $f$. For example, $f$ could be an object classifier, a
segmentation network or an autoencoder.  $f$ maps an image $x \in \RR^{h \times
w \times 3}$ through a sequence of hidden layers to a final output $f(x)$.
Intermediate activations $E(x) \in \RR^{H \times W \times C}$ of a hidden layer
are a task-specific representation of the image $x$. Such hidden
representations convey no meaning to a human and we must transform them into
meaningful representations. We introduce the notation $z = E(x) \in \RR^{H
\cdot W \cdot C}$, \ie $z$ is the $N=H \cdot W \cdot C$ dimensional, flattened
version of the hidden representation to be interpreted. $E$ is the sub-network
of $f$ consisting of all layers of $f$ up to and including the hidden layer
that produces $z$, and the sub-network after this layer will be denoted by
$\tail$, such that $f(x) = \tail \circ E(x)$ as illustrated in Fig.~\ref{fig:overview}.

To turn $z$ into an interpretable representation, we aim to translate the
distributed representation $z$ to a factorized representation
$\tz=(\tz_k)_{k=0}^K \in \RR^N$ where each of the $K+1$ factors $\tz_k \in
\RR^{N_k}$, with $\sum_{k=0}^K N_k = N$, represents an
interpretable concept. The goal of this translation is twofold: On the one
hand, it should enable an analysis of the relationship between data and
internal representations of $f$ in terms of interpretable concepts; this
requires a forward map $T$ from $z$ to $T(z)=\tz$.  On the other
hand, it should enable semantic modifications
on internal representations of $f$; this requires the
inverse of $T$.  With this inverse map, $T^{-1}$, an internal representation
$z$ can be mapped to $\tz$, modified in semantically meaningful ways to obtain
$\tz^*$
(\eg changing a single interpretable concept), and mapped back to an
internal representation of $f$. This way, semantic modifications, $\tz \mapsto
\tz^*$, which were
previously only defined on $\tz$ can be applied to internal
representations via $z\mapsto z^* \defeq T^{-1}(T(z)^*)$. See
Fig.~\ref{fig:cmnistswapping} for an example, where $z$ is modified by
replacing one of its semantic factors $\tz_k$ with that of another image.

\cmnisttransfer %

\paragraph{Disentangling Interpretable Concepts:}
For meaningful analysis and modification, each factor $\tz_k$ must represent
a specific interpretable concept and taken together, $\tz$ should support a
wide range of modifications. Most importantly, it must be possible to
analyze and modify different factors $\tz_k$ independently of each other.
This implies a factorization of their joint density $p(\tz) =
\prod_{k=0}^K p(\tz_k)$. To explore different factors, the
distribution $p(\tz_k)$ of each factor must be easy to sample from to gain
insights into the variability of a factor, and interpolations between two
samples of a factor must be valid samples to analyze changes along a path. We
thus specify each factor to be normally distributed which gives
\begin{equation}
  \label{eq:marginal}
  p(\tz) = \prod_{k=0}^K \Normal(\tz_k \cond 0, \Id)
\end{equation}

\newcommand{\factorlabel}{F}
Without additional constraints, the semantics represented by a factor $\tz_k$
are unspecified. To fix this, we demand that \emph{(i)} each factor $\tz_k$
varies with one and only one interpretable concept and \emph{(ii)} it is
invariant with respect to all other variations.
Thus, let there be training image pairs $(x^a, x^b)$ which specify semantics
through their similarity, \eg image pairs containing animals of the same
species to define the semantic concept of `animal species'.
Each semantic concept $\factorlabel \in \{1, \dots, K\}$ defined by such pairs
shall be represented by the corresponding factor $\tz_\factorlabel$ and we
write $(x^a, x^b) \sim p(x^a, x^b \cond F)$ to emphasize that $(x^a, x^b)$ is a
training pair for factor $\tz_F$.
However, we cannot expect to have examples of image pairs for every semantic
concept relevant in $z$. Still, all factors together, $\tz = (\tz_k)_{k=0}^K$,
must be in one-to-one correspondence with the original representation, \ie
$z=T^{-1}(\tz)$.  Therefore, we introduce $\tz_0$ to act as a residual concept
that captures the remaining variability of $z$ which is missed by the semantic
concepts $\factorlabel=1,\dots,K$.

For a given training pair $(x^a, x^b) \sim p(x^a, x^b \cond F)$, the
corresponding factorized representations, $\tz^a = T(E(x^a))$ and
$\tz^b=T(E(x^b))$, must now \emph{(i)} mirror the semantic similarity of $(x^a,
x^b)$ in its $F$-th factor and \emph{(ii)} be invariant in the remaining
factors. This is expressed by a positive correlation factor $\covar_{ab} \in
(0, 1)$ for the $F$-th factor between pairs,
\begin{equation}
  \label{eq:correlation}
  \tz^b_F \sim \Normal(\tz^b_F \cond \covar_{ab} \tz^a_F, (1-\covar^2_{ab}) \Id)
\end{equation}
and no correlation for the remaining factors between pairs,
\begin{equation}
  \label{eq:nocorrelation}
  \tz^b_k \sim \Normal(\tz^b_k \cond 0, \Id) \quad k\in \{0,\dots,K\} \setminus \{F\}
\end{equation}
To fit this model to data, we utilize the invertibility of $T$ to directly
compute and maximize the likelihood of pairs $(z^a, z^b)=(E(x^a), E(x^b))$. We compute
the likelihood with the absolute value of the Jacobian determinant of $T$,
denoted $\abs{T'(\cdot)}$, as
\begin{align}
    p(z^a, z^b \cond F) 
    &= p(z^a) \; p(z^b \cond z^a, F) \\
    &= \abs{T'(z^a)} \; p\left(T(z^a)\right) \cdot \label{eq:joint:marginal} \\
    &\phantom{{} = {}} \abs{T'(z^b)} \; p\left(T(z^b) \cond T(z^a), F\right) \label{eq:joint:conditional}
\end{align}

To be able to compute the Jacobian determinant efficiently, we follow previous
works \cite{kingma2018glow} and build $T$ based on ActNorm, AffineCoupling and
Shuffling layers as described in more detail in Sec.~\ref{sec:iinarchitecture}
of the supplementary. For training we use the negative log-likelihood as our
loss function. Substituting Eq.~\eqref{eq:marginal} into
Eq.~\eqref{eq:joint:marginal}, Eq.~\eqref{eq:correlation} and
\eqref{eq:nocorrelation} into Eq.~\eqref{eq:joint:conditional}, leads to the
per-example loss $\ell(z^a, z^b \cond F)$,
\begin{align}
  \label{eq:perexloss}
  \ell(z^a, z^b \cond F)
  &=
  \sum_{k=0}^K
  \lVert T(z^a)_k \rVert^2
  - \log \abs{T'(z^a)} \\
  &+
  \sum_{k\neq F}
  \lVert T(z^b)_k \rVert^2
  - \log \abs{T'(z^b)} \\
  &+
  \frac{
    \lVert T(z^b)_F - \covar_{ab} \; T(z^a)_F \rVert^2
  }{1-\covar^2_{ab}}
\end{align}
which is optimized over training pairs $(x^a, x^b)$ for all semantic concepts
$F \in \{1, \dots, K\}$:
\begin{equation}
  \loss = \sum_{F=1}^K \expect_{(x^a, x^b)\sim p(x^a, x^b \cond F)} \;
  \ell(E(x^a), E(x^b) \cond F)
  \label{eq:mainobjective}
\end{equation}

Note that we have described the case where image pairs share at least one
semantic concept, which includes the case where they share more than one
semantic concept. Moreover, our approach is readily applicable in the case
where image pairs differ in a semantic concept. In this case,
Eq.~\eqref{eq:correlation} holds for all factors $\tz^b_k, k\in\{0, \dots, K\}
\setminus \{F\}$ and Eq.~\eqref{eq:nocorrelation} holds for factor $\tz^b_F$.
This case will also be used in the next section, where we discuss the
dimensionality and origin of semantic concepts.

\subsection{Obtaining Semantic Concepts}
\label{sec:obtaining}

\paragraph{Estimating Dimensionality of Factors:}
Semantic concepts differ in complexity and thus also in dimensionality. Given
image pairs $(x^a, x^b)\sim p(x^a, x^b \cond F)$ that define the $F$-th
semantic concept, we must estimate the dimensionality of factor $\tz_F$ that
represents this concept. Due to the invertibility of $T$, the sum of dimensions
of all these factors equals the dimensionality of the original representation.
Thus, semantic concepts captured by the network $E$ require a larger share of
the overall dimensionality than those $E$ is invariant to. 

\pictogram %
\estdim %
\mnistinterpolationscompare %

The similarity of $x^a, x^b$ in the $F$-th semantic concept implies a positive
mutual information between them, which will only be preserved in the latent
representations $E(x^a), E(x^b)$ if the $F$-th semantic concept is captured by
$E$.  Thus, based on the simplifying assumption that components of hidden
representations $E(x^a)_i, E(x^b)_i$ are jointly Gaussian distributed, we
approximate their mutual information with their correlation for each component
$i$.  Summing over all components $i$ yields a relative score $s_F$ that serves
as proxy for the dimensionality of $\tz_F$ in case of training images
$(x^a, x^b) \sim p(x^a, x^b \cond F)$
for concept $F$,
\begin{equation}
  s_F = \sum_i \frac{\covariance{E(x^a)_i, E(x^b)_i}}{
        \sqrt{\variance{E(x^a)_i} \;\variance{E(x^b)_i}}}
    .
\end{equation}
Since correlation is in $[-1,1]$, scores $s_F$ are in $[-N, N]$ for
$N$-dimensional latent representations of $E$. Using the maximum score
$N$ for the residual factor $\tz_0$ ensures that all factors have equal
dimensionality if all semantic concepts are captured by $E$. The dimensionality
$N_F$ of $\tz_F$ is then $N_F = \left\lfloor \frac{\exp{s_F}}{\sum_{k=0}^K
\exp{s_k}} N \right\rfloor$.
Tab.~\ref{tab:estdim} demonstrates the feasibility of predicting
dimensionalities with
this approximation.

\paragraph{Sketch-Based Description of Semantic Concepts:}
Training requires the availability of image pairs that depict changes
in a semantic concept.
Most often, a sufficiently large number of such examples is not easy to obtain. The following describes an approach to help a user specify semantic
concepts effortlessly.

Two sketches are
worth a thousand labels: Instead of labeling thousands of images with
semantic concepts, a user only has to provide two sketches, $y^a$ and $y^b$
which demonstrate a change in a concept. For example, one sketch may contain mostly
round curves and another mostly angular ones as in
Fig.~\ref{fig:pictogram}. We then utilize a style transfer algorithm \cite{park2019arbitrary}
to transform each $x$ from the training set into two new images: $x^a$ and
$x^b$ which are stylized with $y^a$ and $y^b$, respectively. The combinations
$(x, x^a)$, $(x, x^b)$ and $(x^a, x^b)$ serve as
examples for a change in the concept of interest.

\animalswapping %
\dfswapping %

\paragraph{Unsupervised Interpretations:}
Even without examples for changes in semantic factors, our approach can still
produce disentangled factors.
In this case, we minimize
the negative log-likelihood of the marginal distribution of hidden
representations $z=E(x)$:
\begin{equation}
  \lossunc = - \expect_{x} %
  \lVert T(E(x)) \rVert^2
  - \log \abs{T'(E(x))}
\end{equation}
As this leads to independent components in the transformed representation, it
allows users to attribute meaning to this representation after training.
Mapping a linear interpolation in our disentangled space back to $E$'s
representation space leads to a nonlinear interpolation on the data manifold
embedded by $E$ (see Fig.~\ref{fig:mnistinterpolation}). This linear structure
allows to explore the representations using vector arithmetics
\cite{upchurch2017deep}. For example, based on a few examples of images with a
change in a semantic concept, we can find a vector representing this concept as
the mean direction between these images (see Eq.~\eqref{eq:avgdir}). In contrast to previous works, we do
not rely on disentangled latent representations but learn to translate
arbitrary given representations into disentangled ones.

\section{Experiments}
The subsequent experiments use the following datasets: 
\textit{AnimalFaces} \citep{liu2019few},
\textit{DeepFashion} \citep{liu2016deepfashion,Liu_2016},
\textit{CelebA} \citep{liu2015faceattributes}
\textit{MNIST}\citep{lecun1998mnist},
\textit{Cifar10} \citep{krizhevsky2009learning}, and
\textit{FashionMNIST} \citep{xiao2017fashionmnist}.
Moreover, we augment MNIST by randomly coloring its images to provide a benchmark for disentangling experiments (denoted \textit{ColorMNIST}).

\subsection{Interpretation of Autoencoder-Frameworks}
\label{sec:aeint}
Autoencoders learn to reconstruct images from a low-dimensional
latent representation
$z = E(x)$. 
Subsequently, we map $z$ onto interpretable factors to perform semantic
image modification.
Note that $z$ is only obtained using a given network;
our invertible interpretation
network has never seen an image itself.

\paragraph{Disentangling Latent Codes of Autoencoders:}
Now we alter the $\tz_k$ which should in turn modify the $z$ in a semantically meaningful manner.  This tests two
aspects of our translation onto mutually disentangled, interpretable representations: First, if its factors have
been successfully disentangled, swapping factors from different
images should still yield valid representations. Second, if the factors
represent their semantic concepts faithfully, modifying a factor should
alter its corresponding semantic concept.
\celebafid %

To evaluate these aspects, we trained an autoencoder on the AnimalFaces
dataset. As semantic concepts we utilize the animal category and a residual factor.
Fig.~\ref{fig:animalswapping} shows the results of
combining the residual factor of the image on the left with the
animal-class-factor of the image at the top. After decoding, the results depict animals from the class of the image at the top. However, their gaze
direction corresponds to the image on the left. This demonstrates a successful
disentangling of semantic concepts in our interpreted space.

The previous case has confirmed the applicability of our approach to roughly aligned images. We now test it on unaligned images of articulated persons on DeepFashion. Fig.~\ref{fig:dfswapping} presents results for attribute swapping as in the previous experiment. Evidently,
our approach can handle articulated objects and enables pose guided human synthesis.

Finally, we conduct this swapping experiment on ColorMNIST to investigate simultaneous disentangling of multiple factors. Fig.~\ref{fig:cmnistswapping} shows a swapping using an interpretation of three factors: digit type, color, and residual. %

\renewcommand{\bottomfraction}{0.5}
\celebainterpolation %
\celebaclusterinterpolations %

\paragraph{Evaluating the Unsupervised Case:}
To investigate our approach in case of no supervision regarding semantic concepts, we analyze its capability to turn
simple autoencoders into generative models. Because our
interpretations yield normally distributed representations, we can
sample them, translate them back onto the latent space of the autoencoder, and
finally decode them to images,
\begin{align}
  \tz &\sim \Normal(\tz \cond 0, \Id) 
\label{eq:sample} 
,\enspace \enspace
  x = \tail(T^{-1}(\tz))
  .
\end{align}
We employ the standard evaluation protocol for generative models and
measure image quality with Fr\'echet Inception Distance (FID scores).
\cite{dai2019diagnosing} presented an approach to generative
modeling using two Variational Autoencoders and achieved results
competitive with approaches based on GANs.
We follow \cite{dai2019diagnosing}
and use an autoencoder architecture based on %
\cite{lucic2017gans} and train on the same datasets
with losses as in \cite{larsen2015autoencoding}. Tab.~\ref{tab:fidbasic} presents mean and
std of FID scores over three trials with 10K
generated images. We significantly improve over state-of-the-art FID reported in \cite{dai2019diagnosing}.
Our approach can utilize a learned similarity metric similar to GANs, which enables them to produce high-quality images. In contrast to approaches based on GANs, we can rely on an autoencoder and a reconstruction loss. This enables stable training and avoids the mode-collapse problem of GANs, which explains our improvement in FID.
\cmnistresponse %
\umapcmnist %

Besides sampling from the model as described by equation \eqref{eq:sample}, our
approach supports semantic interpolation in the representation
space, since the invertible network constitutes a lossless encoder/decoder
framework. We obtain semantic axes $\tilde{z}^{F
\rightarrow \bar{F}}$ by encoding two sets of images $X^{F}=\{x^{F}\},
X^{\bar{F}}=\{x^{\bar{F}}\}$,
showing examples with an attribute in $X^{F}$ and without that
attribute in $X^{\bar{F}}$. Note that these sets are only required after
training, \ie during test time. $\tz^{F \rightarrow \bar{F}}$ is then obtained
as the average direction between examples of $X_{F}$ and
$X_{\bar{F}}$,
\begin{align}
  \label{eq:avgdir}
  \tz^{F \rightarrow \bar{F}} = 
  \frac{1}{\abs{X^{\bar{F}}}}\sum_{x^{\bar{F}} \in X^{\bar{F}}} x^{\bar{F}} -
  \frac{1}{\abs{X^{F}}}\sum_{x^{F} \in X^{F}} x^{F}
.
\end{align}
Such vector arithmetic depends on a meaningful linear
structure of our interpreted representation space. We illustrate this
structure in Fig.~\ref{fig:mnistinterpolation}. Linear walks in our
interpretable space always result in meaningful decoded images, indicating
that the backtransformed representations lie on the data manifold. In
contrast, decoded images of linear walks in the encoder's hidden
representation space contain ghosting artifacts.  Consequently, our model
can transform nonlinear hidden representations to an interpretable space
with linear structure. Fig.~\ref{fig:celebainterpolation} visualizes
a 2D submanifold on CelebA.

Fig.~\ref{fig:celebacluster} provides an example for an interpolation as described in Eq.~\ref{eq:avgdir} between attributes on the CelebA dataset. We linearly walk along the
\textsl{beardiness} and \textsl{smiling} attributes, increasing the former
and decreasing the latter. 

\subsection{Interpretation of Classifiers}
\label{sec:expclsf}
After interpreting autoencoder architectures %
we now analyze classification networks: 
\emph{(i)} A digit classifier on ColorMNIST (accuracy $ \sim 97 \%$). 
To interpret this network, we extract hidden representations $z \in \mathbb{R}^{64}$ just
before the classification head.
\emph{(ii)} A ResNet-50 classifier \citep{he2016deep} trained on classes of AnimalFaces. 
Hidden representations $z \in \mathbb{R}^{2048}$ are extracted after the fully convolutional layers.
\vspace{-0.5em}

\paragraph{Network Response Analysis:}
We now analyze how class output probabilities change under manipulations in the interpretation space:
First, we train the translator $T$ to disentangle $K$ (plus a residual) distinct
factors $\tilde{z}_k$.
For evaluation we modify a single factor
$\tz_k $ while keeping all others fixed. More precisely, we modify $\tz_k$ by
replacing it with samples drawn from a random walk in a harmonic potential (an
Ornstein-Uhlenbeck process, see Sec.~\ref{sec:ornstein} of the supplementary), starting at $\tz_k$. 
This yields a sequence of modified factors
$(\tilde{z}_k^{(1)}, \tilde{z}_k^{(2)}, \dots, \tilde{z}_k^{(n)})$ when
performing $n$ modification steps. We invert every element in this sequence
back to its hidden representation and apply the classifier.
We analyze the response of the network to each modified factor $k$ through the distribution of the %
logits and class predictions.%

\paragraph{Interpreting Classifiers to Estimate their Invariance:}
Network interpretation also identifies the invariance properties of a learned representation.
Here we evaluate invariances of a digit classifier to color. We learn a translation $T$ to
disentangle \textsl{digit} $\tilde{z}_1$, \textsl{color}
$\tilde{z}_2$, and a \textsl{residual} $\tilde{z}_0$.
Fig.~\ref{fig:cmnistresponse} shows the network response analysis. The distribution of $\log
\text{softmax}$-values and predicted classes is indeed not sensitive to
variations in the factor \textsl{color}, but turns out to be quite
responsive when altering the \textsl{digit} representation.
We additionally show
a UMAP \citep{lel2018umap} 
of the reversed factor manipulations in
Fig.~\ref{fig:umapcmnist}
(in black). Since the entire modification occurs within one cluster, this underlines 
that $T$ found a disentangled
representation 
and that the classifier is almost invariant to color.
Additionally, we employ another 1D-UMAP
dimensionality reduction to each factor seperately and then plot their
pair-wise correlation in Fig.~\ref{fig:cmnistresponse}.
Next, we trained a transformer $T$ to evaluate interpretability in case of the
popular ResNet-50. The analysis of three factors, grayscale value $\tz_1$,
roundness $\tz_2$, and a residual $\tz_0$ reveals an invariance of the classifier
towards grayness but not roundness. More details can be found in
Sec.~\ref{sec:resnet50supp} of the supplementary.

\section{Conclusion}
We have shown that latent representations of black boxes can be translated
to interpretable representations where disentangled factors represent
semantic concepts. We presented an approach to perform this translation
without loss of information. For arbitrary models, we provide the ability to
work with interpretable representations which are equivalent to the ones
used internally by the model. We have shown how this provides a better
understanding of models and data as seen by a model. Invertibility of our
approach enables semantic modifications and we showed how it can be used to
obtain state-of-the-art autoencoder-based generative models.
\blfootnote{This work has been supported in part by the German federal ministry
BMWi within the project ``KI Absicherung'', the German Research Foundation (DFG)
projects 371923335 and 421703927, and a hardware donation from NVIDIA
corporation.}

\FloatBarrier
\clearpage

{\small
\bibliographystyle{ieee_fullname}
\bibliography{ms}
}

\input{appendix}
\end{document}

%% file: appendix.tex
\appendix
\onecolumn
\begin{center}
  \textbf{
  \Large A Disentangling Invertible Interpretation Network for \\
Explaining Latent Representations} \\
-- \\
 \textbf{\large Supplementary Materials} \\
\hspace{1cm}
\end{center}

\section{Implementation Details}
\subsection{Invertible Interpretation Network}
\label{sec:iinarchitecture}
As described in Sec.~\ref{sec:train}, we require $T$ to be an invertible transformation. To achieve this, we use an invertible neural network.
In our implementation, this network is built from three invertible layers:
coupling blocks \cite{dinh2016density}, actnorm
layers \cite{kingma2018glow} and shuffling layers. A sequence of these three layers builds one
invertible block, \cf Fig.~\ref{fig:flowblock}. After passing the input $z$ through multiple blocks, \cf Fig.~\ref{fig:transformerarch}, we split the output $\tz$ into factors $(\tz_k)_{k=0}^K$.
\transformerarch
\flowblock

Each shuffling layer uses a fixed, randomly initialized permutation to
shuffle the channels of its input. Actnorm consists of learnable shift and
scale parameters for each channel, which are initialized to provide
activations with zero mean and unit variance. Coupling blocks equally split their
input $h=(h_1,h_2)$ along its channel dimension and compute
\begin{align}
  \tilde{h}_1 &= h_1 \cdot s_1(h_2) + t_1(h_2) \\
  \tilde{h}_2 &= h_2 \cdot s_2(\tilde{h}_1) + t_2(\tilde{h}_1)
\end{align}
where $s_i$, $t_i$ are fully connected networks.

\paragraph{Definition of notation for network architectures}
To describe the architecture we use in our experiments, we introduce the following notation:
\begin{itemize}
\item Conv2d($c_{in}$, $c_{out}$, $k$, $s$, $p$): A two-dimensional convolution operation, working on $c_{in}$ input channels and producing $c_{out}$ output channels. We use square kernels of size $k$. $s$ denotes the stride, $p$ the amount of padding used.
\item ConvT2d($c_{in}$, $c_{out}$, $k$, $s$, $p$): A two-dimensional transposed convolution operation, working on $c_{in}$ input channels and producing $c_{out}$ output channels. We use square kernels of size $k$. $s$ denotes the stride, $p$ the amount of padding used.
\item Linear($c_{in}$, $c_{out}$): Maps a vector $z_1 \in \mathbb{R}^{c_{in}}$ onto a vector $z_2 \in \mathbb{R}^{c_{out}}$.
\end{itemize}

\subsection{Autoencoder Architecture}
The architecture used for our autoencoder experiments is based on
\cite{lucic2017gans}. We replace batch normalization \cite{ioffe2015batch}
by act normalization \cite{kingma2018glow} to avoid issues caused by
differences in batch statistics during training and testing. Details regarding the architecture can be found in Tab.~\ref{tab:archae}. Furthermore, we
remove the highest fully connected layer which results in a more lightweight
model with less parameters, see Tab.~\ref{tab:numparams} for a comparison.
To implement the adversarial loss, we use the discriminator from
\citep{isola2017image}, operating on patches of input images with a
receptive field of 70 pixels.
\aearchitecture
\numparams

\subsection{Classifier Architecture}
We use two different classifier architectures in our experiments. The first
one uses the same encoder as in the autoencoder, followed by a fully
connected layer, \cf Tab.~\ref{tab:archclf}.
The second one uses the ResNet-50 \citep{he2016deep} architecture. We use
the network up to the last convolutional layer for $E$ and the remaining
network producing the class scores for $\tail$.
\clfarchitecture

\subsection{Details on Sketch Based Description of Semantic Concepts}
Efficient generation of training pairs for semantic concepts involves a
style transfer algorithm as described in Sec.~\ref{sec:obtaining}. For this, we
use the approach of \cite{park2019arbitrary},
with the following set of hyperparameters:
\begin{itemize}
\item content loss: $\lambda_c = 1.0$, 
\item style loss: $\lambda_s = 5.0$, 
\item identity losses: $\lambda_{id,1}=50.0$, $\lambda_{id,2}=1.0$
\end{itemize}
In addition, we use the same discriminator architecture as in our
autoenconder experiments to distinguish real from stylized images. We denote
this loss the \textsl{realism prior} and weight it by $\lambda_g = 1.0$.

Using this setting, the model is trained on AnimalFaces (content) and the
Wikiart ~\citep{karayev2013recognizing} (style) datasets.

\subsection{Training Hyperparameters}
\label{sec:hyperflow}
We train all models using a batch size of $25$ and a learning rate of
$10^{-4}$ for the Adam optimizer \cite{kingma2014adam}. The translation
network $T$ is trained by optimizing Eq.~\ref{eq:mainobjective}, where we
fix $\sigma_{ab}=0.9$ for all experiments. Note that $\sigma_{ab} = 0.0$
corresponds to uncorrelated examples $a$ and $b$, whereas $\sigma_{ab}=1.0$
denotes perfect correlation. Hence, we allow for a small amount of
stochasticity when providing pairs $(a,b)$ to account for low-quality pairs
within the training set.
The hyperparameters for the transformer $T$ are:
\begin{itemize}
\item $n_{flow}$: Number of flow blocks (\cf Fig.~\ref{fig:flowblock}) used to build $T$.
\item $H$: Dimensionality of hidden layers in subnetworks $s_i$ and $t_i$.
\item $D$: Depth of subnetworks $s_i$ and $t_i$.
\end{itemize}
We list all configurations of these hyperparameters used in our experiments
in Tab.~\ref{tab:hyperparamsiin}.
\hyperparamsiin

\newpage
\section{Additional Results}
\label{sec:addres}
\paragraph{Semantic Image Manipulations and Semantic Embeddings}
In the case of autoencoders, our invertible interpretation network enables
semantic image modifications. By transforming the latent representation $z$
of an image $x$ to $\tz$, we can modify semantic concepts of the image: We
modify the factor $\tz_k$ corresponding to the semantic concept, invert the
modified representation back to the latent space of the autoencoder and
finally decode it to the semantically modified image.

In Fig.~\ref{fig:cmnistvid} and \ref{fig:celebavid} we manipulate individual
semantic factors by interpolation on the ColorMNIST and CelebA datasets,
respectively. In both cases, colors of the embeddings represent the
semantics of $\tz_1$, in particular for ColorMNIST, the colors represent the digit class, and for
CelebA, the colors represent the gender. The top shows the interpolation status for
each of the semantic concepts. Next, to the left we display a two-dimensional embedding
of the residual space which illustrates the Gaussian structure of our prior
and its independence with respect to $\tz_1$. To the right we plot a
one-dimensional embedding of $\tz_1$ against a one-dimensional embedding of
$\tz_2$, providing a semantically meaningful two-dimensional embedding. For
ColorMNIST, the observed product structure of this embedding shows the
independence of $\tz_1$ and $\tz_2$. On the other hand, on CelebA we observe
missing data points in the top-left quadrant, demonstrating a lack of
training examples showing women with beards.
\cmnistvid
\celebavid

We then invert the modified representation back to the latent space of the
autoencoder and visualize the resulting representation in a two-dimensional
embedding of the latent space (bottom left). In the animated versions, which
can be found at
\href{https://compvis.github.io/iin/}{https://compvis.github.io/iin/}, we can see
that semantic modifications, which have a simple linear structure in $\tz$, get
mapped to complex paths in $z$ due to the entangled structure of the latent
space.  Finally, the bottom right shows the semantically modified image
$G(T^{-1}(\tz))$.

\paragraph{Partial Sampling of Factors}
Instead of swapping disentangled factors of provided pairs, we can sample a
factor $\tz_k$ while keeping other factors $\tz_i$ fixed. See
Fig.~\ref{fig:dfsampling} for an application on the DeepFashion dataset.
\dfappsampling

\paragraph{Manipulating a Network's Decision by Meaningful Variation of Disentangled Factors}
\label{sec:ornstein}
As described in the main text, we modify a factor $\tz_k$ while keeping all other factors fixed and analyze the response of the classifier by inverting and decoding the code $\tz$.
More precisely, we change each factor by performing a random walk in transformed space, keeping a relation to the input example. To regularize the walk to stay within proximity of the input example, we use a \textsl{Ornstein-Uhlenbeck} process to simulate the walk, \cf Fig.~\ref{fig:brownornstein}. This process can be expressed as a simplified discretized version of a stochastic differential equation:
\begin{equation}
\tz_{k, t+1} = -\gamma \tz_{k,t} + \sigma W_t
,
\end{equation}
where $t$ indexes the random sequence, starting at $\tz_k \equiv \tz_{k,0}$, $W_t \in \mathcal{N}(0, \Id)$ and $\gamma$ and $\sigma$ scalar parameters.
We repeat the analysis done in the main text for a classifier trained on ColorMNIST, see Fig. ~\ref{fig:umapornsteincmnist}.
Again, changing factor \textsl{color} has no effect on the prediction of the classifier (hidden representations stay within the same UMAP cluster), whereas changes in factor \textsl{digit} cause variations in the classifier's prediction.
\brownvsornstein
\umapornsteincmnist

\paragraph{ResNet-50 Classifier Response Analysis}
\label{sec:resnet50supp}
To analyze the expressiveness of our disentangling interpretation approach,
we train an invertible transformation on a ResNet-50 classifier trained to
perform class prediction on AnimalFaces. To this end, we interpret the
effect of three factors: \textsl{greyscale}, \textsl{roundness} (i.e.
softness of contours) and a residual factor.  As can be concluded from the
response analysis (\cf Sec.~\ref{sec:expclsf} and
Fig.~\ref{fig:resnetresponse}), the classifier is not sensitive to changes
in factor \textsl{greyscale}, but does often change its class prediction
when altering the factor \textsl{roundness} (right-hand side of
Fig.~\ref{fig:resnetresponse}, where log-probabilities are plotted). This
suggests that the classifier is (to some degree) relying on the shape of the
contours when predicting the respective class. This behavior is further
confirmed by visualizing the variation within clusters of the classifier's
hidden code by a 2D UMAP embedding in Fig.~\ref{fig:umapresnet}.
\resnetresponse
\umapresnet

\paragraph{Interpretable Representations $\tz$ Improve Sampling-Based Image Synthesis w.r.t. Latents $z$}
Fig.~\ref{fig:celebabothstagesamples},~\ref{fig:cifarbothstagesamples} and
~\ref{fig:fashionmnistbothstagesamples} provide insight into how our
translation network $T$ can map a complex, hidden representation of a given
network onto a interpretable and accessible representation. Here, we compare
decoded samples $x$ when drawing (i) from the prior of the autoencoding
network, i.e. $x=G(z)$ for $z \sim \mathcal{N}(0, \mathbf{1})$ and (ii) from
the prior of the transformer network, i.e. $x=G(T^{-1}(\tz))$ for $\tz \sim
\mathcal{N}(0, \mathbf{1})$. Samples obtained from $\tz$-space yield
structured and more coherent images than samples from the latent space $z$.
These figures also provide a qualitative examples for the quantitative
results in Tab.~\ref{tab:fidbasic} on unconditional image synthesis.

\celebabothstagesamples
\cifarbothstagesamples
\fashionmnistbothstagesamples